\documentclass[preprint,12pt]{elsarticle}

\usepackage{amssymb}
\usepackage{amsmath}
\usepackage{rotating}
\usepackage{hyperref}
\usepackage{array}
\usepackage{multirow}
\usepackage{graphicx}
\usepackage{float}
\usepackage{subfig}
\usepackage{multicol}

\journal{Nuclear Physics B}

\begin{document}

\begin{frontmatter}



\title{Enhanced Urdu Intent Detection with Large Language Models and Prototype-Informed Predictive Pipelines} %% Article title

% \affiliation[label1]{organization={},
% \affiliation[label2]{organization={},

\author[UET,RPTU]{Faiza Hassan\corref{cor2}}
\affiliation[UET]{organization={Department of Electrical Engineering, University of Engineering and Technology},
            addressline={GT Road},
            city={Lahore},
            postcode={54890},
            state={Punjab},
            country={Pakistan}}
\affiliation[RPTU]{organization={Department of Computer Science, Rheinland-Pfälzische Technische Universität Kaiserslautern-Landau (RPTU)},
            addressline={Gottlieb-Daimler-Straße 47},
            city={Kaiserslautern},
            postcode={67663},
            state={Rhineland-Palatinate},
            country={Germany}}
\author[RPTU,DFKI]{Summra Saleem\corref{cor2}}
\affiliation[DFKI]{organization={German Research Center for Artificial Intelligence (DFKI)},    
            addressline={Gottlieb-Daimler-Straße 47},
            city={Kaiserslautern},
            postcode={67663},
            state={Rhineland-Palatinate},
            country={Germany}}
\author[UET]{Kashif Javed}
\author[DFKI,Intell]{Muhammad Nabeel Asim\corref{cor2}}
\affiliation[Intell]{organization={Intelligentx GmbH (intelligentx.com)},    
            % addressline={Gottlieb-Daimler-Straße 47},
            city={Kaiserslautern},
            % postcode={67663},
            % state={Rhineland-Palatinate},
            country={Germany}}
\author[UOG]{Abdur Rehman}
\affiliation[UOG]{organization={Department of Computer Science and Information Technology, University of Gujrat, Jalalpur Jattan Road, Gujrat, 50700, Punjab, Pakistan},    
            addressline={Jalalpur Jattan Road},
            city={Gujrat},
            postcode={50700},
            state={Punjab},
            country={Pakistan}}
\author[RPTU,DFKI,Intell]{Andreas Dengel}

\cortext[cor1]{Corresponding Author. \\ Email Address: \href{mailto:muhammad\_nabeel.asim@dfki.de}{Muhammad\_Nabeel.Asim@dfki.de} (Muhammad Nabeel Asim).}
\cortext[cor2]{These authors contributed equally to this work.}
% \tnotetext[eq]{\textsuperscript{*} These authors contributed equally to this work.}

%%%%%%%%%%%%%%%%%%%%%%%%%%%%%%%%%%%%%%%%%%%%%%%%%%%%%%%%%%%%%%%%%%

%% Abstract
\begin{abstract}
Multifarious intent detection predictors are developed for different languages, including English, Chinese and French, however, the field remains underdeveloped for Urdu, the $10^{\text{th}}$ most spoken language. In the realm of well-known languages, intent detection predictors utilize the strategy of few-shot learning and prediction of unseen classes based on the model training on seen classes. However, Urdu language lacks few-shot strategy based intent detection predictors and traditional predictors are focused on prediction of the same classes which models have seen in the train set. To empower Urdu language specific intent detection, this introduces a unique contrastive learning approach that leverages unlabeled Urdu data to re-train pre-trained language models. This re-training empowers LLMs representation learning for the downstream intent detection task. Finally, it reaps the combined potential of pre-trained LLMs and the prototype-informed attention mechanism to create a comprehensive end-to-end LLMPIA intent detection pipeline. Under the paradigm of proposed predictive pipeline, it explores the potential of 6 distinct language models and 13 distinct similarity computation methods. The proposed framework is evaluated on 2 public benchmark datasets, namely ATIS encompassing 5836 samples and Web Queries having 8519 samples. Across ATIS dataset under 4-way 1 shot and 4-way 5 shot experimental settings LLMPIA achieved 83.28\% and 98.25\% F1-Score  and on Web Queries dataset produced 76.23\% and 84.42\% F1-Score, respectively. In an additional case study on the Web Queries dataset under same classes train and test set settings, LLMPIA outperformed state-of-the-art predictor by 53.55\% F1-Score.
\end{abstract}


\begin{highlights}
\item Introduces a contrastive learning approach to re-train pre-trained language models (PLMs) using unlabeled Urdu data, thus improve their ability to capture semantic nuances.
\item Proposes an end-to-end few-shot learning framework, LLMPIA that combines pre-trained LLMs with a prototype-informed attention mechanism to predict unseen intent classes.
\item Evaluates six pre-trained language models alongside their re-trained versions to measure the impact of contrastive learning within the LLMPIA pipeline.
\item Benchmarks one Urdu-specific and five multilingual PLMs to determine the optimal model choice for Urdu intent detection applications.
\item Investigates thirteen distinct similarity measures to identify the most effective approach for prototype-informed intent detection.
\end{highlights}

%% Keywords

\end{frontmatter}



\section{Introduction}
In the landscape of natural language processing (NLP), intent detection is an essential task to recognize user's needs or purposes from their communication \cite{casanueva2020efficient}. Primarily, intent detection is an integral part of numerous NLP applications such as virtual assistants \cite{baek2024implementation, winkler2024slot}, chatbots \cite{chandrakala2024intent, ouaddi2025dsl, lester2004conversational, laranjo2018conversational}, and automated customer support systems \cite{xu2024retrieval}, etc. However, unlike many traditional NLP tasks such as text classification \cite{asim2021benchmarking, asim2017effect, mehmood2023enml, saleem2023fnreq, ibrahim2021ghs,  saleem2024mlr, li2023logistic, ai2025contrastive, mao2025parameter, riyadi2025benchmarking}, sentiment analysis \cite{mehmood2020precisely, du2022emotion, hellwig2025exploring, bensoltane2025neural, motevalli2025aspect}, hate speech detection \cite{mehmood2024passion}, paraphrasing \cite{wasim2019lexical} and name entity recognition \cite{li2022udbbc, li2025named, yan2025ltner}, intent detection poses significant challenges because of the dynamic nature of human intentions.  While significant advancements have been made in development of intent detection applications for high-resource languages like English \cite{jbene2025intent, lu2025enhancing, yang2025gfidf, yin2025midlm, ferrera2025linguistics, birkmose2025device}, there is a pressing need to extend this capability to Urdu \cite{magueresse2020low}, one of the most widely used languages in South Asia. Around the world, Urdu has over 170 million speakers and is the national language of Pakistan \footnote{https://mdcresearch.net/general-multicultural-resources-for-providers/information-for-providers-who-work-with-families-parents-who-read-chinese/}. 

Similar to other languages, Urdu language-specific intent detection applications are accompanied by significant challenges due to numerous factors such as individuals expressing their intentions in a variety of ways, influenced by their personality, knowledge, experiences and contextual factors \cite{shams2022improving}. A potential challenge is compiling a comprehensive intent detection dataset that covers all possible categories is nearly impossible, leading to the challenge of handling unseen intent categories. In advanced languages, the data annotation problem is addressed by designing predictors capable of identifying samples belonging to unseen categories based on training with samples from known categories \cite{qu2024divide, shams2022improving, shams2019lexical}. In contrast, current Urdu language intent detection predictors only assign intent categories to samples that fall within the categories present in their training dataset. In Urdu language, another major challenge is to generate a comprehensive and semantically rich statistical representation of the text \cite{zhang2022newintent}. While language-specific and multilingual pre-trained language models  \cite{atuhurrazero, cheng2024cyclical, comi2023zero, kathakali2023effect, qu2024divide, tahir2024benchmarking} offer a foundation for Urdu NLP applications, their general training necessitates specialized re-training to enhance performance for specific Urdu language tasks. 
This study aims to design a unique, end-to-end Urdu intent detection predictive pipeline, that addresses the aforementioned challenges through the following key contributions:
\begin{itemize}
    \item With an aim to generate a comprehensive and semantically rich statistical representation of Urdu text, it presents a unique contrastive learning \cite{chuang2020debiased} approach to re-train pre-trained language models (PLMs) \cite{min2023recent} using unlabeled Urdu data. The contrastive learning approach enables the model to learn words representation by distinguishing subtle similarities between text pairs based on their semantic meaning.
    \item It presents a unique framework that encompasses end-to-end few-shot learning strategy based Urdu intent detection predictive pipelines (LLMPIA) that synergistically combines the strengths of pre-trained LLMs with a prototype-informed attention mechanism. LLMPIA is capable of predicting unseen intent classes based on its learning from seen intent classes. 
    \item To assess the effectiveness of the proposed contrastive learning re-training strategy within the LLMPIA pipeline, this study systematically compares the performance of six pre-trained language models against their re-trained versions.
    \item To address the critical question of whether multilingual or Urdu language-specific pre-trained language models are more effective for Urdu intent detection, this research benchmarks the performance of one Urdu-specific and five multilingual models. This comparative analysis provides valuable insights into the optimal model selection for Urdu intent detection application development.
    \item Within the realm of few-shot intent detection, it performs an extensive analysis of thirteen different similarity computation methods, aiming to identify the most optimal similarity measure for developing highly accurate prototype-informed intent detection predictors, providing a solid foundation for future research in few-shot learning for Urdu.
\end{itemize}    

\section{Related Work}
AI-driven intent detection applications landscape has witnessed development of both multilingual and language-specific predictors. Multilingual predictors offer the advantage of broad applicability across various languages but they struggle in capturing the fine-grained syntactic and semantic details that are necessary to identify users intent. In contrast, language-specific predictors are exclusively trained on a particular language and manage to produce better performance by acquiring a deep understanding of that language's specific linguistic characteristics. \\
Existing multilingual intent detection predictors exhibit significant variations in their linguistic scope. For instance, the predictor developed by Atuhurra et al. \cite{atuhurrazero} targets 4 specific languages (Urdu, English,
Japanese, and Swahili) while Cheng et al. \cite{cheng2024cyclical} and Qu et al. \cite{qu2024divide} predictors are developed to handle 9 (Hindi, Turkish, Spanish, German, Japanese, Chinese, Portuguese, French, Thai) and 2 (English, Chinese) languages respectively. Conversely, Comi et al. \cite{comi2023zero} and Xue et al. \cite{xue2021intent}'s predictors are designed to handle 3 languages, German, Italian, English and Thai, Spanish, English concurrently, while Mitra et al.'s model boasts the broadest coverage, processing 51 languages simultaneously. Notably, among these, only the Atuhurra et al. \cite{atuhurrazero} predictor includes Urdu language support. In this predictor development study, the authors explored the zero-shot intent detection capabilities of 3 large language models including GPT-4, Claude 3 Opus, and Gemma. Primarily, authors derived inference based on 6 different prompting approaches namely zero-shot, few-shot Chain-of-Thoughts, zero-shot Chain-of-Thought, Expert-General, Expert-Specific and Multi-Persona. Based on experimental results, authors concluded that substantial morphological complexity inherent in languages like Urdu and Swahili, poses a significant linguistic challenge that results in lower performance outcomes compared to languages English and Japanese. \\
On the other hand, In the realm of AI-driven language-specific intent detection, numerous predictors have been developed for various languages but Urdu language has witnessed development of only 3 predictors \cite{shams2019lexical, shams2022improving, tahir2024benchmarking}. Based on working paradigm these predictors fall into 2 classes 1) traditional \cite{shams2019lexical, shams2022improving} 2) zero-shot \cite{tahir2024benchmarking}. Traditional predictors \cite{shams2019lexical, shams2022improving} utilize the potential of Capsule Neural Netwrok and LSTM to capture complex patterns of Urdu language data. \cite{shams2022improving} investigated the combined potential of Bidirectional LSTM and Capsule Neural Networks for intent detection. While \cite{shams2019lexical} leverages CNN and LSTM with Capsule Neural Networks. These predictors are trained on a predefined set of classes and require extensive labelled data which limits their ability to adapt dymanic staruture of classes. \\
To address the labeld corpus requirement of traditional predictors, \cite{tahir2024benchmarking} employed zero-shot learning strategy \cite{tahir2024benchmarking} with Large Language Models including GPT, Bloomz, LLaMA and Ministral. However, their results indicate relatively low performance due to absence of domain-specific intents. In contrast to traditional or zero shot learning strategies, few-shot approach introduces a shift towards a new direction for intent detection with limited labeled data. It balance the performance and resource efficiency while enhancing accuracy for specific contexts. Specifically, for the Urdu language, this learning technique remains unexplored. Table \ref{literature} presents a comprehensive characteristics of six multilinguagl and 3 Urdu language specific intent detection predictors in terms of dataset, predictor, experimental settings and performance.

%%%%%%%%%%%%%%%%%%%%%%%%%%%%%%%%%%%%%%%%%%%%%%%%%%%%%%%%%%%%%%%%%%%%%%%%%%%%%%
\begin{sidewaystable*}
% \begin{table}
\caption{Literature Review}
\label{literature}
\renewcommand{\arraystretch}{1.5}
\resizebox{1\textwidth}{!}{
\begin{tabular}{|c| c| c| c| c| c| c|}
\hline
\textbf{Author, Year} & \textbf{Dataset} & \textbf{Data Statistics} & \textbf{Representation Learning} & \textbf{Classifier} & \textbf{Experiment Setting} & \textbf{Results} \\ \hline
Qu et al., 2024 \cite{qu2024divide} & \begin{tabular}[c]{@{}c@{}}1. AG news 2. Yahoo answers \\ 3. Yelp reviews 4. Amazon \\ reviews 5. DBpedia 6. MultiUI\end{tabular} & \begin{tabular}[c]{@{}c@{}}No. of Queries, Primary labels: \\ 1: 31,900, 4 2: 1,440,00, 10, \\ 3: 140,000, 5, 4: 730, 000, \\ 5, 5: 3,220,000, 14, 3,468,966, 29\end{tabular} & ERINE & MLP & 10-fold cross validation & \begin{tabular}[c]{@{}c@{}}Accuracy 1. 0.9481 2. 0.8184 \\ 3. 0.7515 4. 0.7116 5. 0.9891 \\ 6. 0.9327\end{tabular} \\ \hline
Atuhurra et al., 2024 \cite{atuhurrazero} & 1. Atuhurra et al. Dataset & \begin{tabular}[c]{@{}c@{}}1. No. of Intents: 6, No. of \\ Queries: 33,812 (8,543 English, \\ 8,543 Japanese, 8,543 Swahili, \\ and 8,543 Urdu)\end{tabular} & GPT-4, Claude 3 Opus & \_ & \_ & Accuracy = $\sim$95 \\ \hline
Cheng et al., 2024 \cite{cheng2024cyclical} & 1. MultiATIS++ 2. MTOP & \begin{tabular}[c]{@{}c@{}}1. MultiATIS++, Language, No of \\ Intents, No of Samples Spanish, German, \\ Chinese, Japanese, Portuguese, 18, 5,871, \\ Hindi, 17, 2,493 Turkish, 17, 1,353 2. MTOP, \\ Languge, No of Samples English, 22,288, \\ German, 18,788, Sapinsh 15,459, \\ French, 16,584, Thai, 15,195, Hindi, 16,131\end{tabular} & BERT & \_ & \begin{tabular}[c]{@{}c@{}}Language, train, validation, test \\ 1. Hindi, 1,440, 160, 893, Turkish, \\ 578, 60, 715, Spanish, 4,488, 490, \\ 893, German, 4,488, 490, 893, \\ Japanese, 4,488, 490, 893, Chinese, \\ 4,488, 490, 893, Portuguese, 4,488, 490, \\ 893, French, 4,488, 490, 893 \\ 2. 70\% train, 10\% validation, 20\% test\end{tabular} & Average Accuracy 1. 93.81 2. 82.24 \\ \hline
Tahir et al., 2024 \cite{tahir2024benchmarking} & 1. Urdu Web Queries & 1. 99,718 instances & GPT-3.5 & \_ & \_ & F1-Score = 0.3 \\ \hline
Comi et al., 2023 \cite{comi2023zero} & \begin{tabular}[c]{@{}c@{}}1. SNLI 2. Banking77-OO S\\ (DE, IT, English)\end{tabular} & \begin{tabular}[c]{@{}c@{}}1. No. of Samples: 570,000 2. No. of \\ Seen Intents: 50, No. of Unseen \\ Intents: 27, No. of Samples: 13,083\end{tabular} & BERT & \_ & \begin{tabular}[c]{@{}c@{}}train, validation, test 1. 550,152 Samples, \\ 10,000, 10,000 2. No. of Seen Intents:\\ 50, No. of Unseen Intents: 27\end{tabular} & \begin{tabular}[c]{@{}c@{}}Accuracy 1. 0.204 2. 0.396 \\ 3. 0.404 4. 0.407\end{tabular} \\ \hline
Mitra et al., 2023 \cite{kathakali2023effect} & MASSIVE & \begin{tabular}[c]{@{}c@{}}No. of Utterances: \textgreater{}51M across \\ 51 languages, No. of Intents: 60\end{tabular} & text-embedding-ada-002 & \_ & \_ & in form of graphs \\ \hline
Shams et al., 2022 \cite{shams2022improving} & \begin{tabular}[c]{@{}c@{}}Shams et al. Dataset \\ (UWQ-22)\end{tabular} & \begin{tabular}[c]{@{}c@{}}8,518 queries, cover 11 domains \\ with 3 intents\end{tabular} & \_ & \begin{tabular}[c]{@{}c@{}}BiLSTM, Capsule \\ Neural Network\end{tabular} & \begin{tabular}[c]{@{}c@{}}80\% train, 10\%\\ development, 10\% testing\end{tabular} & Accuracy = 0.9112 \\ \hline
Xue et al., 2021 \cite{xue2021intent} & \begin{tabular}[c]{@{}c@{}}1. SNIPS-SLU 2. SMP-2018, \\ Facebooks's Multilingual\\(3. EN, 4. TH, 5. SP)\end{tabular} & \begin{tabular}[c]{@{}c@{}}No. of Seen Intents, No. of Unseen \\ Intents, No. of Samples: 1: 5, 2, \\ 1,380 2: 24, 6, 2,460 3: 9, \\ 3, 4,284 4: 9, 3, 8,643 5: 9, 3, 5,353\end{tabular} & Word2Vec, BERT & ReCapsNet & 70\% train, 30\% test & \begin{tabular}[c]{@{}c@{}}Accuracy 1. 0.9752 2. 0.6331 \\ 3. 0.9501 4. 0.8923 5. 08694\end{tabular} \\ \hline
Shams et al., 2019 \cite{shams2019lexical} & 1, ATIS 2. AOL Web Queries & Intents, Queries 1: 26, 5,871 2: 3, 60,000 & n-gram & CNN, C-LSTM & 1. \_ , 2. 80\% Training, 20\% Testing & \begin{tabular}[c]{@{}c@{}}Accuracy 1. ATIS Dataset \\ (CNN): 0.927 2. AOL Web \\ Queries (C-LSTM) 0.641\end{tabular} \\ \hline
\end{tabular}}
% \end{table}
\end{sidewaystable*}
%%%%%%%%%%%%%%%%%%%%%%%%%%%%%%%%%%%%%%%%%%%%%%%%%%%

\section{Research Objective}

In most of the languages, intent detection predictors leverage few-shot learning strategies to predict unseen classes by generalizing from a limited number of examples in seen classes. This approach enhances the model's adaptability and robustness in real-world applications. However, for the Urdu language, there is a noticeable gap in the development of intent detection models that incorporate few-shot learning capabilities, and existing predictors are capable of predicting only classes present in the training data. This restricts their ability to generalize to new, unseen intents, thereby limiting their practical applicability in dynamic settings. To bridge this gap, the proposed LLMPIA predictor paradigm for few-shot intent detection focuses on acquiring generalizable knowledge from a small set of known classes, which can then be transferred to identify new, unseen categories during deployment. We denote the entire text category as:
\[
C_{\text{all}} = \{C_{\text{labeled, known}}, C_{\text{unlabeled, unknown}}\}
\]

To prepare a pre-trained language model (PLM) that is adaptable to the current domain, we conduct contrasitive learning based pre-training using unlabeled text. Let 
\[
D_{\text{unlabeled}} = \{x_i \mid y_i \in C_{\text{all}}\}
\]
represent the unlabeled text dataset utilized during this phase. 
% Here, $f(\cdot)$ denotes the PLM, and $\theta$ represents its parameters.

In the second stage of the few-shot intent detection process, we utilize the standard process known as the $N$-way $K$-shot paradigm. This process helps in generating prototype representations for different intent categories. The entire intent detection dataset is divided into three distinct sets:

\begin{itemize}
    \item \textbf{Training Set} ($D_{\text{train}}$): This set consists of labeled text from a known category, denoted as $C_{\text{train}} \in C_{\text{labeled, known}}$.
    \item \textbf{Validation Set} ($D_{\text{val}}$): This set includes unlabeled text from a category represented as $C_{\text{val}} \in C_{\text{unlabeled, unknown}}$.
    \item \textbf{Test Set} ($D_{\text{test}}$): Similar to the validation set, this contains unlabeled text from a category denoted as $C_{\text{test}} \in C_{\text{unlabeled, unknown}}$.
\end{itemize}

The categories in $C_{\text{train}}$, $C_{\text{val}}$, and $C_{\text{test}}$ are mutually exclusive, meaning there is no overlap between them:
\[
C_{\text{train}} \cap C_{\text{val}} \cap C_{\text{test}} = \emptyset.
\]

\subsection{LLMPIA-Training Phase}

During the LLMPIA-training phase:
\begin{itemize}
    \item \textbf{Support Set}: From $D_{\text{train}}$, we create a collection of $N \times K$ labeled samples $(x_s, y_s)$, where $x_s$ is the text and $y_s$ is the label. Each of the $N$ categories has $K$ examples.
    \item \textbf{Query Set}: A collection of $N \times Q$ unlabeled samples $(x_q, y_q)$, where $Q$ is the number of queries per category. Here, $x_q$ indicates the text input, and $y_q$ signifies its associated label.
\end{itemize}

\subsection{LLMPIA-Validation and LLMPIA-Testing Phases}

For both the LLMPIA-validation and LLMPIA-testing phases:
\begin{itemize}
    \item We similarly extract text from the validation set $D_{\text{val}}$ and test set $D_{\text{test}}$ to construct support and query sets.
\end{itemize}

This structured approach enables efficient learning and adaptation of models to classify intents with minimal labeled examples, while leveraging broader knowledge from unlabeled data.
\section{Proposed Framework}

This section comprehensively describes the proposed large language model-driven prototype-informed predictor for Urdu language intent detection. The framework primarily leverages six different LLMs and employs two different approaches: (1) LLMRCL: Large Language Models Enhanced Representations with Contrastive Learning based re-training followed by (2) prototype-informed attention mechanism based intent detection predictive pipeline LLMPIA. The proposed framework leverages 6 distinct pre-trained language models including \href{https://huggingface.co/google-bert/bert-base-multilingual-cased}{BERT-base-104-languages} \cite{wu2020all}, \href{https://huggingface.co/distilbert/distilbert-base-multilingual-cased}{DistilBERT-base-104-languages}, \href{https://huggingface.co/microsoft/mdeberta-v3-base}{DeBERTa-base-100-languages}, \href{https://huggingface.co/Urduhack/roberta-Urdu-small}{RoBERTa-small-Urdu}, \href{https://huggingface.co/google/muril-base-cased}{MuRIL-base-17-languages} \cite{khanuja2021muril} and \href{https://huggingface.co/Geotrend/bert-base-15lang-cased}{BERT-base-15-languages}. This section begins with a high-level overview of 6 distinct pre-trained language models used. The subsequent sub-sections then provide an in-depth explanation of domain-specific self-supervised learning and the prototype-informed attention.
Table \ref{pre-trained} illustrates 6 distinct pre-trained LLMs in terms of used training data, language of training data, attention head and scale.
These models vary significantly in specialization ranging from broad multilingual models like \href{https://huggingface.co/google-bert/bert-base-multilingual-cased}{BERT-base-104-languages} , 
\href{https://huggingface.co/distilbert/distilbert-base-multilingual-cased}{DistilBERT-base-104-languages} and \href{https://huggingface.co/microsoft/mdeberta-v3-base}{DeBERTa-base-100-languages}, which are trained over 100 languages, to more focused models such as \href{https://huggingface.co/Urduhack/roberta-Urdu-small}{RoBERTa-small-Urdu} which is specifically trained for the Urdu language. Additionally, models like \href{https://huggingface.co/google/muril-base-cased}{MuRIL-base-17-languages} and \href{https://huggingface.co/Geotrend/bert-base-15lang-cased}{BERT-base-15-languages} emphasize regional and domain-specific adaptations, incorporating languages commonly spoken in South Asia, including Urdu. Furthermore, all models except DistilBERT consist of 12 encoder layers, whereas DistilBERT has a more compact architecture comprising only 6 encoder layers. These models exhibit differences in parameters, attention mechanisms, and scale, reflecting trade-offs between computational efficiency and contextual understanding. The diverse array of pre-trained LLMs ensures robust evaluation of the proposed framework's performance by leveraging both general-purpose and Urdu-specialized representations for intent detection.
%%%%%%%%%%%%%%%%%%%%%%%%%%%%%%%%%%%%%%%%%%%%%%%%%%%%%%%

\renewcommand{\arraystretch}{1.05}\label{tab:pre-trained-model}
\begin{table}[ht]
% \begin{longtable}{|p{2.3cm}|p{2.3cm}|p{2.4cm}|p{3.5cm}|p{3.5cm}|}
\caption{A comprehensive overview of pre-trained language models used in this study in terms of their key attributes }
\label{pre-trained}
\renewcommand{\arraystretch}{1.25}
\resizebox{1.0\textwidth}{!}{
\begin{tabular}{c c c c c}
\hline
\textbf{Language Model} & \textbf{Training Data} &\textbf{ Data Statistics} & \textbf{Language} &\textbf{ Attention Heads \& Parameters} \\ \hline
\href{https://huggingface.co/google-bert/bert-base-multilingual-cased}{BERT-base-104-languages} & \begin{tabular}[c]{@{}c@{}}Top 104 languages \\ wikipedia data\end{tabular} & \_ & 104 languages & \begin{tabular}[c]{@{}c@{}}Attention Head = 12, \\ Total Parameters = 110 M\end{tabular} \\ \hline
\begin{tabular}[c]{@{}c@{}} \href{https://huggingface.co/google/muril-base-cased }{MuRIL-base-17-languages} \\(BERT) \end{tabular} & \begin{tabular}[c]{@{}c@{}}17 languages wikipedia \\ data and Common \\ Crawl OSCAR Corpus\end{tabular} & \begin{tabular}[c]{@{}c@{}}Total: 3,030.27 GB \\ Urdu: 2312.08 MB\end{tabular} & \begin{tabular}[c]{@{}c@{}}Assamese, Bengali,English \\ Gujarati, Hindi, Kannada, \\ Kashmiri, Malayalam, Marathi, \\ Nepali, Oriya, Punjabi, Sanskrit, \\ Sindhi, Tamil, Telugu, Urdu\end{tabular} & \begin{tabular}[c]{@{}c@{}}Attention Head = 12, \\ Total Parameters = 16B\end{tabular} \\ \hline
\begin{tabular}[c]{@{}c@{}} \href{https://huggingface.co/Geotrend/bert-base-15lang-cased}{BERT-base-15-languages} \end{tabular} & \begin{tabular}[c]{@{}c@{}} 15 languages\\ wikipedia data \end{tabular} & 1.5M & \begin{tabular}[c]{@{}c@{}}English, French, Spanish, \\ German, Chinese, Arabic, \\ Russian, Vietnamese, Greek, \\ Bulgarian, Thai, Turkish, \\ Hindi, Urdu, Swahili\end{tabular} & \begin{tabular}[c]{@{}c@{}}Attention Head = 12, \\ Total Parameters = 141M\end{tabular} \\ \hline
\href{https://huggingface.co/distilbert/distilbert-base-multilingual-cased}{DistilBERT-base-104-languages} & \begin{tabular}[c]{@{}c@{}} 104 languages \\ wikipedia data \end{tabular} & \_ & 104 languages & \begin{tabular}[c]{@{}l@{}}Attention Head = 12, \\ Total Parameters = 134M\end{tabular} \\ \hline
\href{https://huggingface.co/microsoft/mdeberta-v3-base}{DeBERTa-base-100-languages} & \begin{tabular}[c]{@{}c@{}} CC-100 Multilingual \\ Dataset \end{tabular} & Total: 2.5T Urdu: 884M & 100 languages & \begin{tabular}[c]{@{}l@{}}Attention Head = 12, \\ Total Parameters = 22M\end{tabular} \\ \hline
\href{https://huggingface.co/Urduhack/roberta-Urdu-small}{RoBERTa-small-Urdu} & Urdu news corpus & 50k & Urdu & \begin{tabular}[c]{@{}c@{}}Attention Head = 12, \\ Total Parameters = 125M\end{tabular} \\ \hline
\end{tabular}}
\end{table}
%%%%%%%%%%%%%%%%%%%%%%%%%%%%%%%%%%%%%%%%%%%%%%%%%%%%%%%%%
\subsection{LLMRCL: Large Language Models Enhanced Representations with Contrastive Learning based re-training}
\begin{figure}[htbp]
    \centering
    \includegraphics[width=1.0\linewidth]{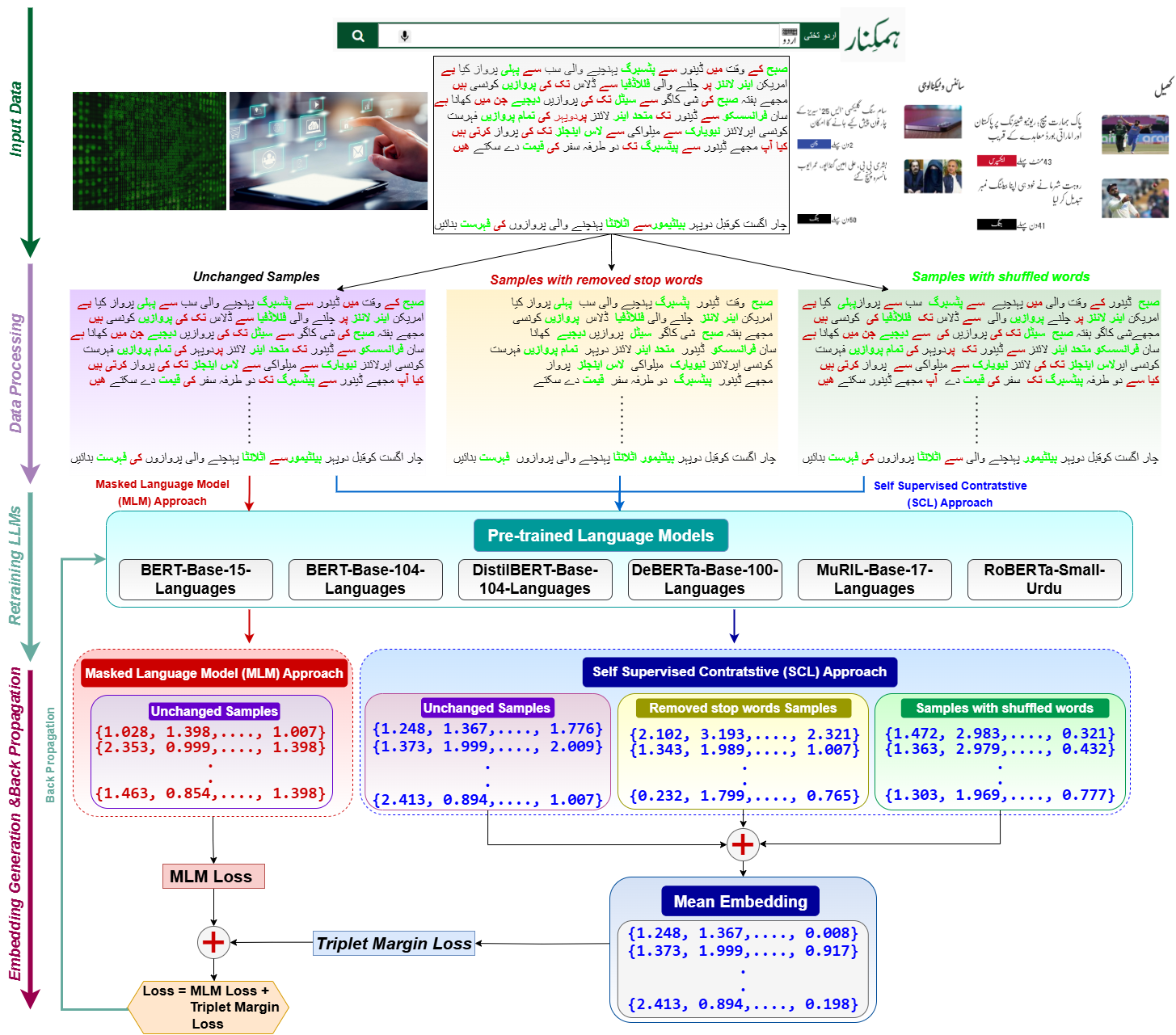}
    \caption{Graphical illustration of Key Steps of Large Language Models Enhanced Representations with Contrastive Learning based re-training: 1) Masked Language Modelling, 2) Self-Supervised Contrastive Learning }
    \label{LLMCRL_1}
\end{figure}

Large Language Models Enhanced Representations with Contrastive Learning based re-training (LLMRCL) empowers pre-trained language models by effectively integrating prior knowledge with domain-specific information to enable PLM to effectively capture semantic relation of the current domain. Moreover, LLMCRL enhances the quality of sentence representation which are pivotal for intent detection. Mainly, LLMCRL encompasses two key tasks: a masked language modelling (MLM) task that facilitates the learning of past domain knowledge and a self-supervised contrastive learning (SCL) that is comparable to textual entailment. Key steps involved in LLMCRL process are graphically shown in Figure \ref{LLMCRL_1} and briefly summarized in following sub-sections.

\textbf{Masked Language Modelling:} MLM self-supervised learning technique \cite{radford2019language, devlin2019bert, liu2019roberta} randomly choses a subset of words from the input sentence of pre-trained language model to be masked at a predefined masking rate. The model is then trained to predict these masked words based on context words. The process enhances word association capability of models which compels them to learn in-depth semantic relation of text. In this study, 25\% of the tokens are randomly selected for prediction. Among these selected tokens, 80\% are substituted with the ``[MASK]” token, 10\% are replaced with random tokens and the remaining 10\% are left unchanged. \\
\textbf{Self-Supervised Contrastive Learning:} Similar to textual entailment, self-supervised contrastive learning \cite{jaiswal2020survey, kim2020adversarial, albelwi2022survey} task enables the model to independently learn to distinguish the semantic relation between pair of texts. The primary goal of model is to compare the semantic differences between positive or negative examples relative to anchor text in order to determine the discriminative representation that capture the intent attribute. To create positive and negative examples pairs different pre-processing strategies are applied over existing domain data. The model is then trained with a contrastive loss to optimize its representation learning.  

\begin{equation}
\mathcal{L}_{\text{LLMCRL}} = \mathcal{L}_{\text{mlm}} \left( D_{\text{unlabeled}} ; \theta \right) + \mathcal{L}_{\text{scl}} \left( D_{\text{unlabeled}} ; \theta \right)
\label{LLMCRL}
\end{equation}
\subsection{Prototype Informed Attention Model}
\begin{figure}[htbp]
    \centering
    \includegraphics[width=1.0\linewidth]{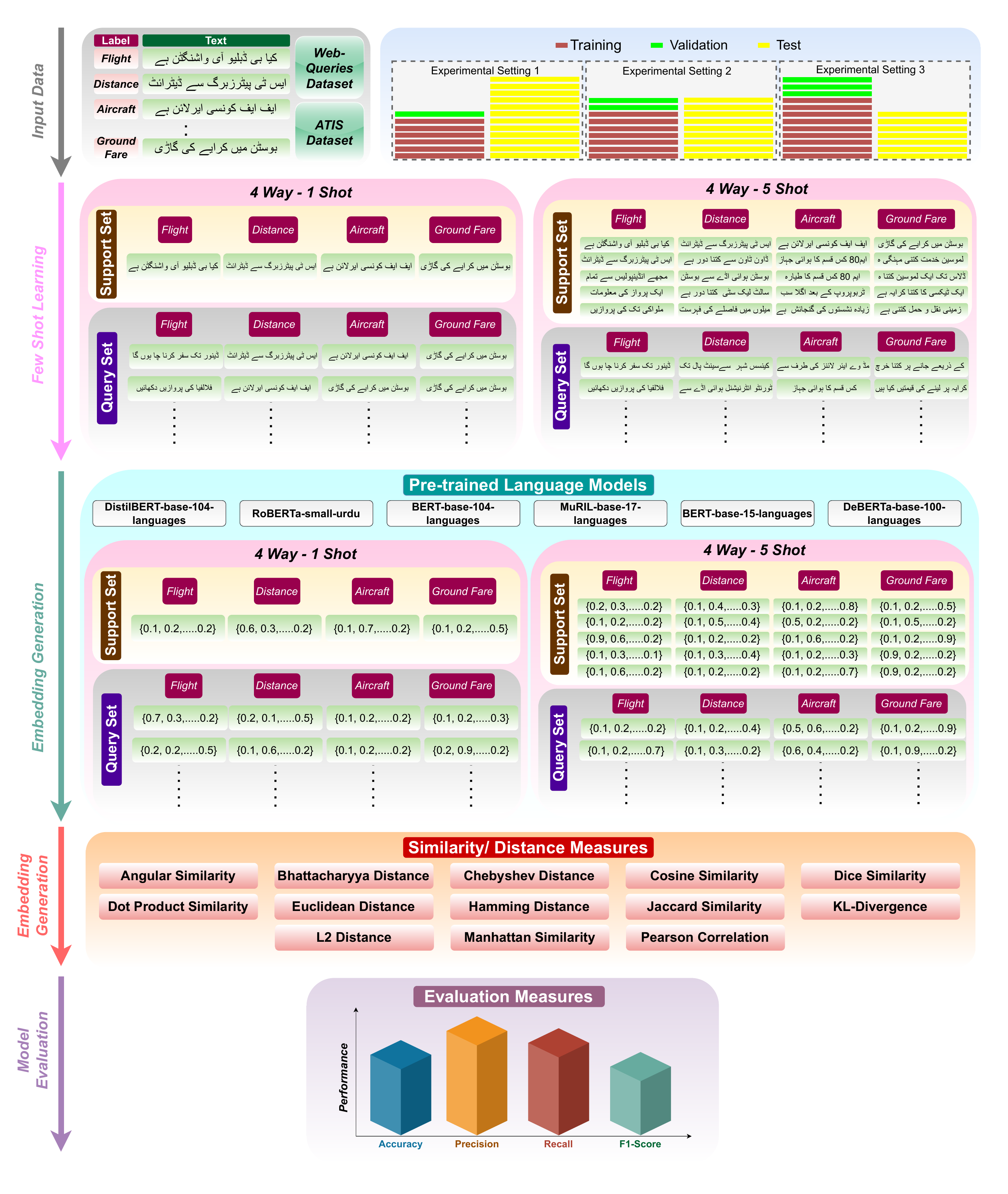}
    \caption{Graphical illustration of the key steps in the Prototype Informed Attention Approach for Urdu intent detection}
    \label{PIA}
\end{figure}

The Prototype-Informed Attention (PIA) \cite{snell2017} paradigm is developed on top of Siamese architecture to facilitate complex feature interactions by generating accurate class prototype representations. PIA promotes consistent feature representation learning between prototype and query set representations through parameter sharing. This method captures a broader range of feature information which allows more nuanced comparisons between prototypes and unknown text queries at the sentence level. The key components of PIA model includes, embedding layer, feature interaction attention, prototype-informed layer, adaptive layer, unsupervised contrastive regularisation and intent detection metric learning. Figure \ref{PIA} graphically illustrates crucial components of PIA approach.

\textbf{Embedding layer:} Six language models pre-trained using LLMCRL approach serves as the embedding layer. These models are enriched with domain-specific knowledge which further enhances the intent detection performance. The episode task \(\{(x^{s}, y^{s}), \; (x^{q}, y^{q})\} \in D_{\text{train}}\) is specifically extracted from the training datasets (as shown in Equation \ref{embedding-layer}) and fed into models enhanced with LLMCRL for embedding representation.
\begin{equation}
XS, \; XQ = f_{\text{LLMCRL}}(x^{s}, \; x^{q})
\label{embedding-layer}
\end{equation}
In Equation \ref{embedding-layer}, \(XS \in \mathbb{R}^{NK \times L_{\text{seq}} \times D_{h}}, \quad XQ \in \mathbb{R}^{NQ \times L_{\text{seq}} \times D_{h}}\) represent the embedding representations containing \(N\times K\) and \(N \times Q\) support set texts, respectively; \(L_{\text{seq}}\) and \(D_{h}\) represent the feature vector representation's dimension and the sequence length of word embedding, respectively.

\textbf{Features interaction attention (FIAT) layer:} FIAT enhances the model's ability to capture complex interactions between different features of an input sentence. FIAT layers focuses on individual features as well as interactions between them to provide richer representations. Sentence-level feature interactions assists to identify key intent keywords by learning how text features belonging to same intent class interact. This enables model to generate sentence-level representations for unseen texts to ensure these representations align with prototype features using the shared parameters of the Feature Interaction Attention Layer (FIAT).

In order to generate prototypes, FIAT combines embedding vectors (\(XS\) and \(XQ\)) of the same category before changing the shape of \(XS \in \mathbb{R}^{N \times K \times L_{\text{seq}} \times D_{h}}\). Then FIAT generates inner-sentence level \(I_\text{sent}\) and inner-class level \(I_\text{cla}\) feature attention information using Equation \ref{FIAT-2}. 
\begin{equation}
I = \text{LN} \left( \text{MaxPool} \left( \text{softmax} \left( \frac{Q \cdot K^{T}}{\sqrt{d_{h}}} \right) \cdot V \right) \right)
\label{FIAT-2}
\end{equation}

%%%%%%%%%%%%%%%%%%%%%%%%%%%%%%%%%%%%%%%%%%%%%%%%%%%
where: 

\(\mathbf{Q}, \mathbf{K}, \mathbf{V}\) are learnable linear transformations applied to \(\mathbf{XS}\), either at the sentence level (\(\mathbf{Q}_{\text{sent}}, \mathbf{K}_{\text{sent}}, \mathbf{V}_{\text{sent}}\)) or class-level (\(\mathbf{Q}_{\text{cla}}, \mathbf{K}_{\text{cla}}, \mathbf{V}_{\text{cla}}\)).

The dimensions of \(\mathbf{Q}, \mathbf{K}, \mathbf{V}\) adjust accordingly:

For sentence-level attention:$
\mathbf{Q}, \mathbf{K}, \mathbf{V} \in \mathbb{R}^{N \times \text{head} \times K \times L_{\text{seq}} \times d_h}$

For class-level attention:$
\mathbf{Q}, \mathbf{K}, \mathbf{V} \in \mathbb{R}^{N \times \text{head} \times d_h \times K \times L_{\text{seq}}}$

%%%%%%%%%%%%%%%%%%%%%%%%%%%%%%5
Lastly, FIAT incorporates the acquired inner-sentence level and inner-class level feature information into a prototype representation denoted as textproto1, which is the hamrmonic mean between \(I_{\text{sentence}}\) and \(I_{\text{class}}\).

To determine the inner-sentence level feature information for the feature extraction of an unknown text representation \(XQ\), FIAT utilizes Equation \ref{FIAT-4}.
\begin{equation}
    X_Q^{\text{sentence}} = \text{LN}\left( \text{sof} \, \text{tmax}\left( Q_{X_Q} \cdot K_{X_Q}^T \sqrt{d_h} \right) \cdot V_{X_Q} \right)
    \label{FIAT-4}
\end{equation}
In Equation \ref{FIAT-4}, \(\{ Q_{X_Q}, K_{X_Q}, V_{X_Q} \} \in \mathbb{R}^{N_Q \times \text{head} \times L_{\text{seq}} \times d_h}\), \(X_Q^{\text{sentence}} \in \mathbb{R}^{N_Q \times L_{\text{seq}} \times D_h}\). The query matrix \(Q = \{ (Q_{\text{sent}}, Q_{\text{cla}}), Q_{XQ} \}\), key matrix \(K = \left\{ \left( K_{\text{sent}}, \, K_{\text{cla}} \right) \right\}\) and value matrix \(V = \{ (V_{\text{sent}}, V_{\text{cla}}), V_{XQ} \}\) used above, are obtained
by the same linear transformation, as shown in Equation \ref{q-k-v}.
\begin{equation}
    \begin{bmatrix}
    Q \\
    K \\
    V
    \end{bmatrix}
    =
    \begin{bmatrix}
    w_q \\
    w_k \\
    w_v
    \end{bmatrix}
    \cdot x +
    \begin{bmatrix}
    b_q \\
    b_k \\
    b_v
    \end{bmatrix}
    \label{q-k-v}
\end{equation}

In Equation \ref{q-k-v}, \(x={XS,XQ}\); \(w_q , w_k , w_v\) denote \(query\) linear transformation, \(key\) linear transformation, and \(value\) linear transformation, respectively. Additionally, the bias terms of each linear transformation are indicated by \(b_q\), \(b_k\), and \(b_v\).

\textbf{Prototype informed Layer (PI Layer):} The purpose of the PI Layer is to improve and polish prototype representation. The idea is to gather each prototype's internal sentences' input. Weighted to a certain prototype representation after being compared to the prototype, which is the significance of every sentence in the prototype. The method of calculation is as follows: Equation \ref{PI layer}.
\begin{equation}
    \text{proto}_2 = \text{proto}_1 \cdot \text{softmax}\left( \tanh\left( w_{PI} \cdot \text{proto}_1 + b_{PI} \right) \right)
    \label{PI layer}
\end{equation}
In Equation \ref{PI layer}, \(proto_2 \in \mathbb{R}^{N \times D_h}, \quad w_{PI}\) and \(b_{PI}\) represent the weight parameters and
bias terms of the PI layer, and \(tanh(.)\) represents the tangent hyperbolic function. On the other hand, the computation module does not require a query set for unknown text.

\textbf{Adaptive Layer (Ad Layer).} Adaptive layer leverages shared parameters of Siamese network to establish the mapping relationship of the underlying features.  The prototype representation and the unknown text query set representation are mapped into a common space by the Ad layer. This improves representation of prototype and query set and aid the model to comprehend the connections between the prototype and query. The Ad layer, composed of a feedforward neural network performs its computations based formula provided in Equation \ref{AD layer}.
\begin{equation}
    proto, XQ_p=AD(proto_2, XQ_{sent})
    \label{AD layer}
\end{equation}
In Equation \ref{AD layer}, \(proto \in \mathbb{R}^{N \times D_h}\) denotes the final prototype representation and \(XQ_p \in \mathbb{R}^{NQ \times D_h}\) represents the unknown text.

\textbf{Unsupervised Contrastive Regularization.} Few-shot intent detection methods frequently produce very fitting examples. In order to stop PIA from overfitting in few-shot learning, unsupervised contrastive regularisation is employed. Learning separable representations among different prototypes and varying unknown texts aids to prevent overfitting. To acquire \(proto'\) and \(XQ'\) same computation strategy of \(proto\) and the \(XQ_p\) is followed ,(where \(XS'\) and \(XQ'\) are gotten by \(XS\) and \(XQ\) with \(dropout\) pre-processing, and \(XQ'\) get \(proto'\) and \(XQ_{p}'\)). Equation \ref{regularization} provide mathematical expression for unsupervised contrastive regularisation computation approach.
 \begin{equation}
    \mathcal{L} = -\frac{1}{N} \sum_{i=1}^{N} \log \left( \frac{\exp\left(\frac{z_i \cdot z_i^{+}}{\tau}\right)}{\exp\left(\frac{z_i \cdot z_i^{+}}{\tau}\right) + \sum_{j \neq i} \exp\left(\frac{z_i \cdot z_j^{-}}{\tau}\right)} \right)
    \label{regularization}
\end{equation}
where $z\_i$ is a generic term representing a feature vector (either $proto_i$ (for computing ucl1) or $XQ_{p,i}$ (for computing ucl2)). \\
$z_i^{+}$ and $z_j^{-}$ are the corresponding positive and negative counterparts of $z_i$ and $z_j$, $\tau$ is the temperature hyperparameter while N is the total number of elements, just as before.

\textbf{Intent Detection Metric Learning.} Intent detection metric learning module comprises of a metric function with a temperature coefficient and a cross-entropy loss function to maximise the learning of PIA. To determine the similarity between the prototype representation and the unknown text representation, cosine similarity function is used with a temperature coefficient. Equation \ref{cosine similarity} illustrates mathematical formulation to compute cosine similarity.
\begin{equation}
    \text{sim}{\cos} = \frac{\cos\left(XQ_{p}, \ \text{proto}\right)}{t}
     \label{cosine similarity}  
\end{equation}
In Equation \ref{cosine similarity}, \(simcos \in \mathbb{R}^{NQ \times N}\) represents the similarity score. The prototype with the highest similarity score is chosen for intent identification in order to reflect the unknown text's intent category. The model's performance is directly influenced by the size of the temperature coefficient, denoted by \(t\). Once the similarity score has been determined, cross-entropy loss value between the similarity score and the true label determine the model's intent prediction error as shown in Equation \ref{cross-entropy}
\begin{equation}
    \mathcal{L}_{ce} = -\frac{1}{NQ} \sum_{i=1}^{NQ} \sum_{j=1}^{N} y_{ij}^q \cdot \log\left(\text{simcos}_{ij}\right)
    \label{cross-entropy}
\end{equation}
In conclusion, $\mathcal{L}_{\text{total}} = \mathcal{L}_{ce} + \mathcal{L}_{ucl1} + \mathcal{L}_{ucl2}$ is the learning purpose of the entire PIA.

\subsection{Training Process}
%%%%%%%%%%%%%%%%%%%%%%%%%%%%%%%%%%%%%%%%%%%%%%%%%%%%%%%

During the re-training of models in the LLMCRL, we combined train, test and validation sets of each dataset. For the MLM task, words were randomly masked and the masked language modeling error is computed. For the SCL task, stop words are removed and the words within sentences are randomly shuffled to help the model effectively capture semantic relationships between words. The combined error from both tasks is utilized to optimize the model. In PIA mechanism, prototype representations of the support set are generated and used to predict the intent of unlabeled queries. For the few-shot learning strategy, we adopted the concept of episodes during the training phase, as it is a common approach. However, due to its inability to reliably capture real-world performance, episodes are not used in the testing phase.

%
%%%%%%%%%%%%%%%%%%%%%%%%%%%%%%%%%%%%%%%%%%%%%%%%%%%%%%%%%%%%%%%%%%%%%%%%%%%%%%%%%%%
\section{Datasets and Evaluation Measures}
This section provides a detailed overview of the datasets used to evaluate the proposed framework, along with the evaluation measures used to assess its performance.
\subsection{Datasets}

\begin{figure}[htbp]
    \centering
    % First Image
    \begin{minipage}{0.45\textwidth}
        \centering
        \includegraphics[width=\textwidth]{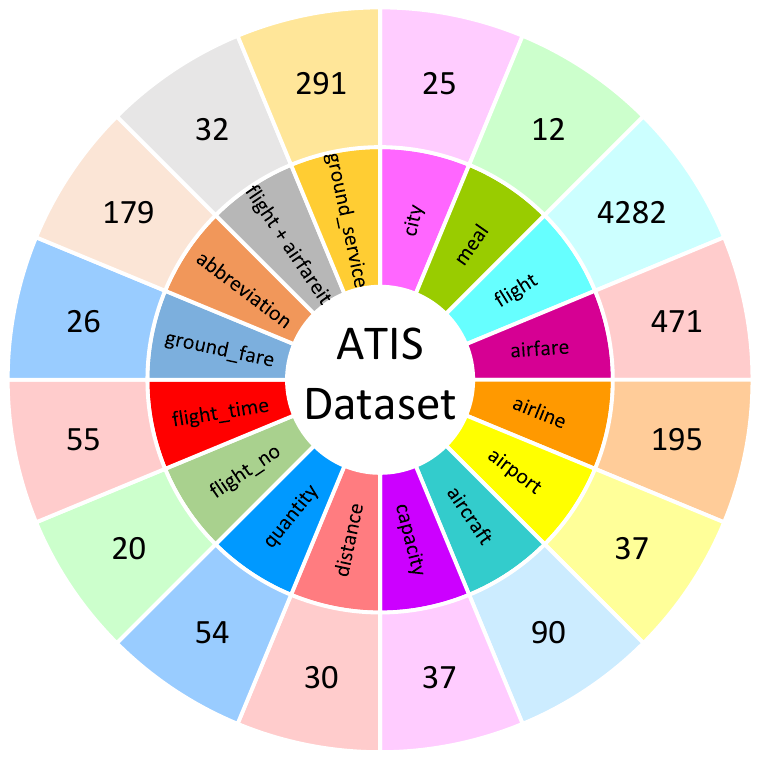}
        \par (a) ATIS Dataset % Subfigure label
    \end{minipage}
    \hfill
    %Second Image
    \begin{minipage}{0.45\textwidth}
        \centering
        \includegraphics[width=\textwidth]{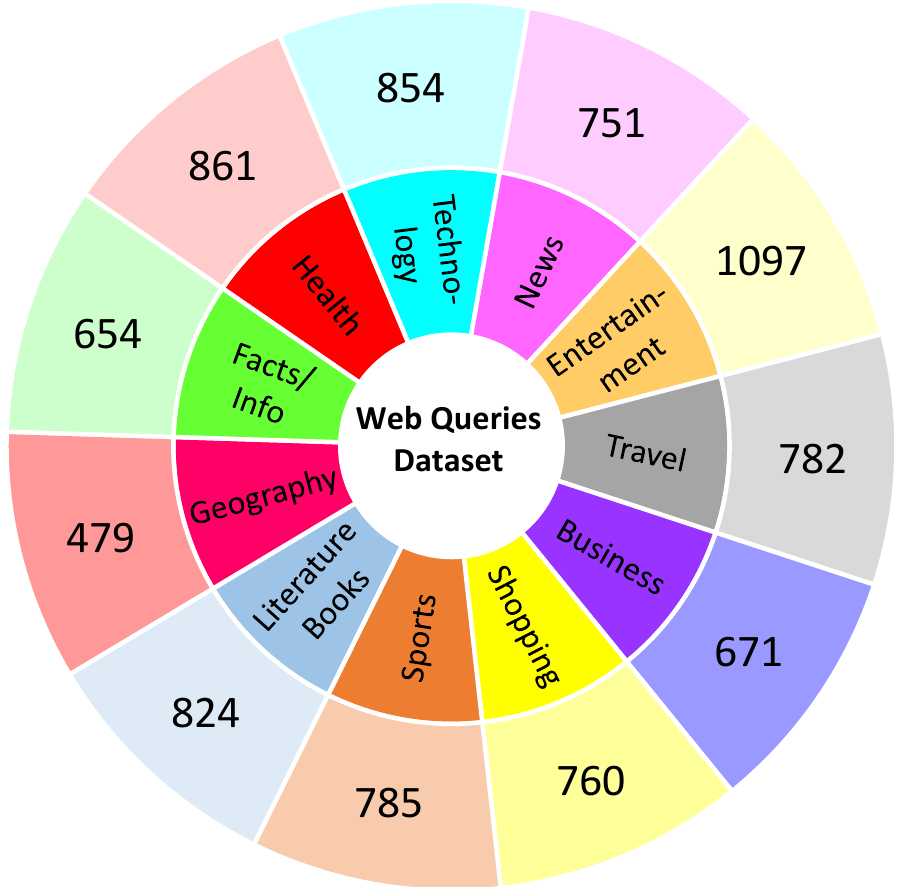}
        \par (b) Web Queries Dataset% Subfigure label
    \end{minipage}
    \caption{Graphical overview of dataset statistics for Urdu intent detection}
    \label{graphical overview}
\end{figure}

To facilitate development of Urdu Intent detection applications, two separate research articles, authored by Shams et al.,\cite{shams2019lexical} and \cite{shams2022improving} have introduced 2 distinct datasets namely Urdu Web Queries and ATIS (Airline Travel Information System). The development of ATIS dataset relied on a resource-constrained approach in which authors translated existing English language intent detection datasets into Urdu using Google's Translation Toolkit. Notably, the labels associated with each sample were not translated and retained their original English format within the Urdu dataset. Moreover, to ensure accuracy, native Urdu speakers reviewed and corrected translations where the original meaning was distorted. This dataset is annotated with 26 flight-related intents, such as airline, flight, abbreviation, ground-services, airfare, quantity, abbreviate, and airporte. The original corpus contains 899 unique words, and this vocabulary size was preserved after translating the dataset into Urdu. Each user query is consistently formatted by placing the query between a BOS (Beginning of Statement) and EOS (End of Statement) tag, followed by the corresponding Intent label. Classes with six or fewer queries excluded from dataset as such small sample sizes are not suitable for few-shot intent detection. The final dataset contains 5836 queries distributed across 16 classes. \\
Second dataset named Web Queries \cite{shams2022improving} is a pioneering collection of native Urdu search queries. This dataset was compiled using search records from Humkinar, a localized Urdu search engine. It contains 11 domains which cover 3 intents with 8519 queries. Generally, domain refers to the broader context, category, or subject area of a conversation, while intents refers to the specific goal or purpose behind a user’s query or statement. It reflects what the user wants to achieve within a domain. However, Shams et al., \cite{shams2022improving} proposed intents as broader category while domains as narrow concept. Traditional intent detection predictors are evaluated on this dataset, but their primarily focus on predicting intent classes that they have seen in the training set. In contrast, our predictor aims to identify unseen intent classes by leveraging knowledge from the training set or previously observed classes. This evaluation approach is more effective for datasets with a larger number of classes. To fulfill this criterion, we use domain names as intent labels instead of intent classes. In this way, dataset encompases 11 class labels. Additionally, to compare our predictor’s performance with existing models, we also evaluate it on this dataset with three intent class labels. However, in this case, rather than distinguishing between seen and unseen classes, we include all three classes in both the training and test sets. Figure \ref{graphical overview} illustrates a statistical overview of both ATIS and Web Queries datasets.

\subsection{Evaluation Measures}

Following the evaluation criteria of existing intent detection predictors \cite{qu2024divide, shams2022improving, shams2019lexical}, we utilized four evaluation measures to assess performance of distinct predictive pipelines of proposed framework. These measures include accuracy, weighted precision, weighted recall and weighted F1-Score \cite{asim2022adh, asim2022rmlocnet, summra2021supervised, stricker2022circnet}. Equation \ref{standard_evaluation_measures} provides a mathematical expression to compute the aforementioned evaluation measures.

\begin{equation}\label{standard_evaluation_measures}
 \begin{aligned} 
\text{Accuracy} &= \frac{TP + TN}{TP + TN + FP + FN} \\
\text{Weighted Precision} &= \sum_{i=1}^{n} \frac{|\text{Class}_i|}{\text{Total Instances}} \cdot \text{Precision}_i \\
\text{Weighted Recall} &= \sum_{i=1}^{n} \frac{|\text{Class}_i|}{\text{Total Instances}} \cdot \text{Recall}_i \\
\text{Weighted F1-Score} &= \frac{2 \cdot \text{Weighted Precision} \cdot \text{Weighted Recall}}{\text{Weighted Precision} + \text{Weighted Recall}}
\end{aligned}
\end{equation}
In Equation \ref{standard_evaluation_measures}, $n$ denotes number of classes, $\text{Class}_i$ represents the number of instances in the $i$-th class. $\text{Total Instances}$ represents the total number of instances across all classes. $\text{Precision}_i$ and $\text{Recall}_i$, denotes precision and recall for the $i$-th class.
\section{Experimental Setup and Results}
The proposed framework is developed on top of 5 different APIs namely; scikit-learn \footnote{https://scikit-learn.org/}, numpy \footnote{https://numpy.org/}, math \footnote{https://
docs.python.org/3/library/math.html}, pandas \footnote{https://pandas.pydata.org/} and pytorch \footnote{https://pytorch.org/}. Performance of predictive pipelines is evaluated using pre-trained and re-trained versions of language models. Both types of predictive pipelines' performance is evaluated under two distinct few-shot learning settings, namely 4-way 1-shot setting utilized one training example per of the 4 intent classes, and 5 examples for the 5-shot setting. To ensure a fair performance comparison across the proposed framework's predictive pipelines, we performed an extensive hyperparameter search for each predictive pipeline. Table \ref{hyper_ATIS} and \ref{hyper_Web} illustrate hyperparameters search space and selected optimal values for both approaches (1) Large Language Models Enhanced Representations with Contrastive Learning based re-training (LLMCRL) followed by (2) prototype-informed attention machanism (PIA). During the training process, the predictive pipelines are periodically evaluated on the validation set after every 100 learning episodes. Early stopping employed to select each predictive pipeline, optimal hyper-parameters based on aforementioned evaluation strategy.
%%%%%%%%%%%%%%%%%%%%%%%%%%%%%%
\begin{table}[htbp]
%%%%%%%%%%%%%%%%%%%%%%%%%%%%%%%%%%%%%%%%%%
\centering
\caption{Hyper-parameters space and distinct predictive pipelines selected optimal values for LLMCRL and PIA over ATIS dataset}
\label{hyper_ATIS}
\renewcommand{\arraystretch}{1.5}
\resizebox{1.0\textwidth}{!}{ % Adjusted width to avoid overflow
\begin{tabular}{|c|c|c|c|c|c|c|c|}
\hline
\textbf{Hyper-parameter} & \textbf{Search Space} & \textbf{BERT-base-15-languages} & \textbf{BERT-base-104-languages} & \textbf{DistilBERT-base-104-languages} & \textbf{DeBERTa-base-100-languages} & \textbf{MuRIL-base-17-languages} & \textbf{RoBERTa-small-Urdu} \\ \hline
\multicolumn{8}{|c|}{\textbf{Domain Specific Self-Supervised Learning (LLMCRL)}} \\ \hline
Batch Size & \{8, 32, 64, 128\} & 64 & 64 & 64 & 64 & 64 & 64 \\ \hline
Learning Rate & \begin{tabular}[c]{@{}c@{}} $\text{1e}^{-3}$, $\text{1e}^{-4}$, $\text{1e}^{-5}$, $\text{1e}^{-6}$, \\ $\text{2e}^{-3}$, $\text{2e}^{-4}$, $\text{2e}^{-5}$, $\text{2e}^{-6}$ \end{tabular} & $\text{1e}^{-5}$ & $\text{1e}^{-6}$ & $\text{1e}^{-5}$ & $\text{1e}^{-5}$ & $\text{1e}^{-5}$ & $\text{1e}^{-5}$ \\ \hline
Epochs & \{10, 30, 50, 100, 200, 500\} & 50 & 50 & 30 & 200 & 10 & 100 \\ \hline
\multicolumn{8}{|c|}{\textbf{Prototype Informed Attention Model (PIA)}} \\ \hline
Learning Rate & \begin{tabular}[c]{@{}c@{}} $\text{1e}^{-3}$, $\text{1e}^{-4}$, $\text{1e}^{-5}$, $\text{1e}^{-6}$, \\ $\text{2e}^{-3}$, $\text{2e}^{-4}$, $\text{2e}^{-5}$, $\text{2e}^{-6}$ \end{tabular} & $\text{1e}^{-4}$ & $\text{1e}^{-6}$ & $\text{1e}^{-4}$ & $\text{1e}^{-5}$ & $\text{1e}^{-5}$ & $\text{1e}^{-5}$ \\ \hline
Dropout Rate & \{0.05, 0.1, 0.2, 0.3, 0.4, 0.5\} & 0.1 & 0.1 & 0.1 & 0.1 & 0.1 & 0.1 \\ \hline
Hidden Size & \{100, 200, 300, 400, 500\} & 300 & 300 & 300 & 300 & 300 & 300 \\ \hline
\end{tabular}
}
\end{table}
%%%%%%%%%%%%%%%%%%%%%%%%

\begin{table}[htbp]
\caption{Grid search space along with optimal values for LLMCRL and PIA over Web Queries dataset}
\label{hyper_Web}
\renewcommand{\arraystretch}{1.5}
\resizebox{1.0\textwidth}{!}{
\begin{tabular}{|cccccccc|}
\hline
\multicolumn{1}{|c|}{\textbf{Hyper-parameter}} & \multicolumn{1}{c|}{\textbf{Search Space}} & \multicolumn{1}{c|}{\textbf{\begin{tabular}[c]{@{}c@{}}BERT-base-\\ 15-languages\end{tabular}}} & \multicolumn{1}{c|}{\textbf{\begin{tabular}[c]{@{}c@{}}BERT-base-\\ 104-languages\end{tabular}}} & \multicolumn{1}{c|}{\textbf{\begin{tabular}[c]{@{}c@{}}DistilBERT-base-\\ 104-languages\end{tabular}}} & \multicolumn{1}{c|}{\textbf{\begin{tabular}[c]{@{}c@{}}DeBERTa-base-\\ 100-languages\end{tabular}}} & \multicolumn{1}{c|}{\textbf{\begin{tabular}[c]{@{}c@{}}MuRIL-base-\\ 17-languages\end{tabular}}} & \textbf{\begin{tabular}[c]{@{}c@{}}RoBERTa-\\ small-Urdu\end{tabular}} \\ \hline
\multicolumn{8}{|c|}{\textbf{Large Language Models Enhanced Representations with Contrastive Learning based re-training (LLMCRL):}} \\ \hline
\multicolumn{1}{|c|}{Batch Size} & \multicolumn{1}{c|}{8, 32, 64, 128} & \multicolumn{1}{c|}{64} & \multicolumn{1}{c|}{64} & \multicolumn{1}{c|}{64} & \multicolumn{1}{c|}{64} & \multicolumn{1}{c|}{64} & 64 \\ \hline
\multicolumn{1}{|c|}{Learning Rate} & \multicolumn{1}{c|}{\begin{tabular}[c]{@{}c@{}}$1e^{-3}, 1e^{-4}, 1e^{-5}, 1e^{-6},$ \\ $2e^{-3}, 2e^{-4}, 2e^{-5}, 2e^{-6}$\end{tabular}} & \multicolumn{1}{c|}{$1e^{-5}$} & \multicolumn{1}{c|}{$1e^{-6}$} & \multicolumn{1}{c|}{$1e^{-6}$} & \multicolumn{1}{c|}{$1e^{-5}$} & \multicolumn{1}{c|}{$1e^{-6}$} & $1e^{-6}$ \\ \hline
\multicolumn{1}{|c|}{Epochs} & \multicolumn{1}{c|}{10, 30, 50, 100, 200, 500} & \multicolumn{1}{c|}{50} & \multicolumn{1}{c|}{50} & \multicolumn{1}{c|}{50} & \multicolumn{1}{c|}{200} & \multicolumn{1}{c|}{50} & 200 \\ \hline
\multicolumn{8}{|c|}{\textbf{Prototype Informed Attention Model (PIA):}} \\ \hline
\multicolumn{1}{|c|}{Learning Rate} & \multicolumn{1}{c|}{\begin{tabular}[c]{@{}c@{}}$1e^{-3}, 1e^{-4}, 1e^{-5}, 1e^{-6},$ \\ $2e^{-3}, 2e^{-4}, 2e^{-5}, 2e^{-6}$\end{tabular}} & \multicolumn{1}{c|}{$1e^{-5}$} & \multicolumn{1}{c|}{$1e^{-6}$} & \multicolumn{1}{c|}{$1e^{-5}$} & \multicolumn{1}{c|}{$1e^{-5}$} & \multicolumn{1}{c|}{$1e^{-5}$} & $1e^{-5}$ \\ \hline
\multicolumn{1}{|c|}{Dropout Rate} & \multicolumn{1}{c|}{0.05, 0.1, 0.2, 0.3, 0.4, 0.5} & \multicolumn{1}{c|}{0.1} & \multicolumn{1}{c|}{0.1} & \multicolumn{1}{c|}{0.1} & \multicolumn{1}{c|}{0.1} & \multicolumn{1}{c|}{0.1} & 0.1 \\ \hline
\multicolumn{1}{|c|}{Hidden Size} & \multicolumn{1}{c|}{100, 200, 300, 400, 500} & \multicolumn{1}{c|}{300} & \multicolumn{1}{c|}{300} & \multicolumn{1}{c|}{300} & \multicolumn{1}{c|}{300} & \multicolumn{1}{c|}{300} & 300 \\ \hline
\end{tabular} }
\end{table}

%%%%%%%%%%%%%%%%%%%%%%%%%%%%%%%%%%%%%%%%%%%%%%%%%%%%%%%%%%%%
Following subsections provides an in-depth performance comparison of proposed framework predictive pipelines by taking pre-trained and re-trained language models. Furthermore, under the paradigm of re-trained language models based few-shot learning strategy-based predictive pipelines; it explores the potential of 13 distinct similarity. Lastly, it presents proposed predictor performance comparison with existing predictors \cite{shams2022improving}.

\subsection{Language Models re-training impact on Predictive Pipelines Performance}\label{ATIS-dataset-results-table}
%%%%%%%%%%%%%%%%%%%%%%%%%%%%%%%%%
\begin{table}[htbp]
\caption{Accuracy and F1-Score of 4-way 1-shot and 4-way 5-shot over ATIS dataset}
\label{ATIS}
\renewcommand{\arraystretch}{1.5}
\resizebox{1.0\textwidth}{!}{
\begin{tabular}{|l|l|l|ll|ll|ll|ll|ll|ll|}
\hline
\multirow{2}{*}{\textbf{\begin{tabular}[c]{@{}c@{}}Seen \\ Classes\end{tabular}}} & \multirow{2}{*}{\textbf{\begin{tabular}[c]{@{}c@{}}Experimental \\ Setting\end{tabular}}} & \multirow{2}{*}{\textbf{Model}} & \multicolumn{2}{c|}{\textbf{\begin{tabular}[c]{@{}c@{}}BERT-base-15- \\ languages\end{tabular}}} & \multicolumn{2}{c|}{\textbf{\begin{tabular}[c]{@{}c@{}}BERT-base-104- \\ languages\end{tabular}}} & \multicolumn{2}{c|}{\textbf{\begin{tabular}[c]{@{}c@{}}DistilBERT-base-104- \\ languages\end{tabular}}} & \multicolumn{2}{c|}{\textbf{\begin{tabular}[c]{@{}c@{}}DeBERTa-base-100- \\ languages\end{tabular}}} & \multicolumn{2}{c|}{\textbf{\begin{tabular}[c]{@{}c@{}}MuRIL-base-17- \\ languages\end{tabular}}} & \multicolumn{2}{c|}{\textbf{\begin{tabular}[c]{@{}c@{}}RoBERTa-Urdu- \\ small\end{tabular}}} \\ \cline{4-15} 
 &  &  & \multicolumn{1}{l|}{\textbf{Accuracy}} & \textbf{F1-Score} & \multicolumn{1}{l|}{\textbf{Accuracy}} & \textbf{F1-Score} & \multicolumn{1}{l|}{\textbf{Accuracy}} & \textbf{F1-Score} & \multicolumn{1}{l|}{\textbf{Accuracy}} & \textbf{F1-Score} & \multicolumn{1}{l|}{\textbf{Accuracy}} & \textbf{F1-Score} & \multicolumn{1}{l|}{\textbf{Accuracy}} & \textbf{F1-Score} \\ \hline
\multirow{4}{*}{25\%} & \multirow{2}{*}{\begin{tabular}[c]{@{}c@{}}4-way \\ 1-shot\end{tabular}} & Pre-train & \multicolumn{1}{l|}{\textbf{0.5593}} & \textbf{0.6090} & \multicolumn{1}{l|}{0.5464} & 0.5890 & \multicolumn{1}{l|}{0.5255} & 0.5487 & \multicolumn{1}{l|}{0.2797} & 0.3301 & \multicolumn{1}{l|}{0.2926} & 0.3067 & \multicolumn{1}{l|}{0.4136} & 0.4566 \\ \cline{3-15} 
 &  & re-train & \multicolumn{1}{l|}{\textbf{0.5887}} & \textbf{0.6181} & \multicolumn{1}{l|}{0.5817} & 0.6088 & \multicolumn{1}{l|}{0.5711} & 0.6083 & \multicolumn{1}{l|}{0.4795} & 0.4911 & \multicolumn{1}{l|}{0.5504} & 0.5661 & \multicolumn{1}{l|}{0.5652} & 0.5920 \\ \cline{2-15} 
 & \multirow{2}{*}{\begin{tabular}[c]{@{}c@{}}4-way \\ 5-shot\end{tabular}} & Pre-train & \multicolumn{1}{l|}{0.5716} & 0.6123 & \multicolumn{1}{l|}{\textbf{0.6227}} & \textbf{0.6569} & \multicolumn{1}{l|}{0.5206} & 0.5744 & \multicolumn{1}{l|}{0.3512} & 0.3846 & \multicolumn{1}{l|}{0.3587} & 0.3801 & \multicolumn{1}{l|}{0.5106} & 0.5619 \\ \cline{3-15} 
 &  & re-train & \multicolumn{1}{l|}{\textbf{0.6924}} & \textbf{0.7222} & \multicolumn{1}{l|}{0.6899} & 0.7084 & \multicolumn{1}{l|}{0.6413} & 0.6711 & \multicolumn{1}{l|}{0.4700} & 0.5018 & \multicolumn{1}{l|}{0.5405} & 0.5826 & \multicolumn{1}{l|}{0.5567} & 0.5982 \\ \hline
\multirow{4}{*}{50\%} & \multirow{2}{*}{\begin{tabular}[c]{@{}c@{}}4-way \\ 1-shot\end{tabular}} & Pre-train & \multicolumn{1}{l|}{0.3737} & 0.4281 & \multicolumn{1}{l|}{\textbf{0.3782}} & \textbf{0.4288} & \multicolumn{1}{l|}{0.2829} & 0.3370 & \multicolumn{1}{l|}{0.2466} & 0.3033 & \multicolumn{1}{l|}{0.3359} & 0.3445 & \multicolumn{1}{l|}{0.3238} & 0.3179 \\ \cline{3-15} 
 &  & re-train & \multicolumn{1}{l|}{0.4478} & 0.4708 & \multicolumn{1}{l|}{0.4342} & 0.4975 & \multicolumn{1}{l|}{0.3389} & 0.3506 & \multicolumn{1}{l|}{0.4130} & 0.4652 & \multicolumn{1}{l|}{\textbf{0.6263}} & \textbf{0.6652} & \multicolumn{1}{l|}{0.3707} & 0.3600 \\ \cline{2-15} 
 & \multirow{2}{*}{\begin{tabular}[c]{@{}c@{}}4-way \\ 5-shot\end{tabular}} & Pre-train & \multicolumn{1}{l|}{0.5564} & 0.6227 & \multicolumn{1}{l|}{\textbf{0.6025}} & \textbf{0.6629} & \multicolumn{1}{l|}{0.5962} & 0.6502 & \multicolumn{1}{l|}{0.3052} & 0.3463 & \multicolumn{1}{l|}{0.4563} & 0.5075 & \multicolumn{1}{l|}{0.5135} & 0.5572 \\ \cline{3-15} 
 &  & re-train & \multicolumn{1}{l|}{0.7345} & 0.7815 & \multicolumn{1}{l|}{0.7234} & 0.7693 & \multicolumn{1}{l|}{0.6343} & 0.6905 & \multicolumn{1}{l|}{0.4467} & 0.5004 & \multicolumn{1}{l|}{\textbf{0.8092}} & \textbf{0.8175} & \multicolumn{1}{l|}{0.7901} & 0.8161 \\ \hline
\multirow{4}{*}{75\%} & \multirow{2}{*}{\begin{tabular}[c]{@{}c@{}}4-way \\ 1-shot\end{tabular}} & Pre-train & \multicolumn{1}{l|}{0.6686} & 0.7094 & \multicolumn{1}{l|}{0.7294} & 0.7675 & \multicolumn{1}{l|}{0.6804} & 0.7271 & \multicolumn{1}{l|}{0.4717} & 0.5328 & \multicolumn{1}{l|}{0.7392} & 0.7488 & \multicolumn{1}{l|}{\textbf{0.7412}} & \textbf{0.7700} \\ \cline{3-15} 
 &  & re-train & \multicolumn{1}{l|}{0.7882} & 0.8234 & \multicolumn{1}{l|}{0.8529} & 0.8675 & \multicolumn{1}{l|}{0.7745} & 0.8100 & \multicolumn{1}{l|}{0.8563} & 0.8638 & \multicolumn{1}{l|}{\textbf{0.9529}} & 0\textbf{0.9581} & \multicolumn{1}{l|}{0.8804} & 0.8879 \\ \cline{2-15} 
 & \multirow{2}{*}{\begin{tabular}[c]{@{}c@{}}4-way \\ 5-shot\end{tabular}} & Pre-train & \multicolumn{1}{l|}{0.8016} & 0.8369 & \multicolumn{1}{l|}{0.8320} & 0.8640 & \multicolumn{1}{l|}{0.7348} & 0.7815 & \multicolumn{1}{l|}{0.6020} & 0.6521 & \multicolumn{1}{l|}{0.7955} & 0.8328 & \multicolumn{1}{l|}{\textbf{0.8806}} & \textbf{0.9015} \\ \cline{3-15} 
 &  & re-train & \multicolumn{1}{l|}{0.9271} & 0.9344 & \multicolumn{1}{l|}{0.9332} & 0.9371 & \multicolumn{1}{l|}{0.7814} & 0.8114 & \multicolumn{1}{l|}{0.8569} & 0.8757 & \multicolumn{1}{l|}{\textbf{0.9818}} & \textbf{0.9825} & \multicolumn{1}{l|}{0.9150} & 0.9251 \\ \hline
\end{tabular}}
\end{table}
%%%%%%%%%%%%%%%%%%%%%%%%%%%%%%%%%

\normalsize
This section illustrates the performance of purposed prototype-informed Urdu intent detection predictive pipelines under two distinct variants of language models: pre-trained and re-trained. Table \ref{ATIS} presents the performance of 12 predictive pipelines (6 pre-trained models + 6 re-trained models) across ATIS dataset under 2 experimental settings: 4-way 1-shot and 4-way 5-shot, and 3 distinct data splits—25\% seen / 75\% unseen, 50\% seen / 50\% unseen, and 75\% seen / 25\% unseen.
In Table \ref{ATIS}, a thorough examination of pre-trained language model-based predictive pipelines, reveals that performance is significantly influenced by architecture, and capability to learn informative features. Specifically, BERT-base-15-languages pre-trained model based predictive pipeline outperforms others in 4-way 1-shot setting with a 25\% seen data split. On the other hand, BERT-base-104-languages model demonstrates superior performance in both settings including 5 shot of 50\% seen data split and 1 shot with a 25\% seen data split. Primarily, the superior performance of BERT-base-104-languages pre-trained model based predictive pipeline in most cases is due to its extensive multilingual training, which equips it with the ability to generalize across a wide variety of linguistic patterns. Finally, with a 75\% seen data split, under both settings (1 shot and 5 shot), RoBERTa-small-Urdu model based predictive pipeline stands out as the top-performing pre-trained language model. Its specialization in Urdu-specific data allows it to capture the linguistic and semantic pattrens of the target language more effectively particularly when a large proportion of the data is seen during training. However, lightweight models (DeBERTa-base-100-languages and DistilBERT-base-104), underperom acorss all experimental settings. Their smaller architectures and reduced pre-training limit their ability to capture the complexity of intent detection for low-resource language.
Moreover, Table \ref{ATIS} illustrates that re-training significantly boosted the performance of all models, particularly. MuRIL-base-17-languages showed an impressive performance gains at higher data splits with an average accuracy gain of around 25\%. This analysis indicates re-training overcomes the limitations of initial pre-training and comprehensively capture the semantic relation of specific low resource language. Furthermore, 4-way 5-shot consistently outperforms 4-way 1-shot task across all pre-trained and re-trained predictive pipelines. The additional labeled examples in the 5-shot setting provide richer contextual information, which aids models to better generalize.
The size of the data split had a profound impact on model performance, with larger splits consistently leading to better results. With 25\% of the data, all models struggled to achieve competitive results, particularly in the pre-trained state. However, as the data split increased to 50\% and 75\%, substantial improvements are observed, particularly for the re-trained models. For instance, BERT-base-104-languages achieved an F1-Score of 0.7084 in the 4-way 5-shot task with 25\% data split, which improved to 0.9371 with 75\% data split. Similarly, MuRIL-base-17-languages exhibited exceptional performance gains, reaching an F1-Score of 0.9581 in the 75\% split for the 4-way 5-shot setting. This trend highlights the critical role of data availability in enabling models to better understand and generalize in low-resource scenarios, particularly when re-training is applied.
%%%%%%%%%%%%%%%%%%%%%%%%%%%%%%%%%%%%%%%%%%%%%%%%%%%%%%%%%%%%%
\begin{table}[htbp]
\caption{Accuracy and F1-Score of 4-way 1-shot and 4-way 5-shot over Web Queries dataset}
\label{Web_Queries}
\renewcommand{\arraystretch}{1.5}
\resizebox{1.0\textwidth}{!}{
\begin{tabular}{|l|l|l|ll|ll|ll|ll|ll|ll|}
\hline
\multirow{2}{*}{\textbf{\begin{tabular}[c]{@{}c@{}} Seen \\ Classes\end{tabular}}} & \multirow{2}{*}{\textbf{\begin{tabular}[c]{@{}c@{}}Experimental \\ Setting\end{tabular}}} & \multirow{2}{*}{\textbf{Model}} & \multicolumn{2}{c|}{\textbf{\begin{tabular}[c]{@{}c@{}}BERT-base-15- \\ languages\end{tabular}}} & \multicolumn{2}{c|}{\textbf{\begin{tabular}[c]{@{}c@{}}BERT-base-104- \\ languages\end{tabular}}} & \multicolumn{2}{c|}{\textbf{\begin{tabular}[c]{@{}c@{}}DistilBERT-base-104- \\ languages\end{tabular}}} & \multicolumn{2}{c|}{\textbf{\begin{tabular}[c]{@{}c@{}}DeBERTa-base-100- \\ languages\end{tabular}}} & \multicolumn{2}{c|}{\textbf{\begin{tabular}[c]{@{}c@{}}MuRIL-base-17- \\ languages\end{tabular}}} & \multicolumn{2}{c|}{\textbf{\begin{tabular}[c]{@{}c@{}}RoBERTa-small- \\ Urdu\end{tabular}}} \\ \cline{4-15} 
 &  &  & \multicolumn{1}{l|}{\textbf{Accuracy}} & \textbf{F1-Score} & \multicolumn{1}{l|}{\textbf{Accuracy}} & \textbf{F1-Score} & \multicolumn{1}{l|}{\textbf{Accuracy}} & \textbf{F1-Score} & \multicolumn{1}{l|}{\textbf{Accuracy}} & \textbf{F1-Score} & \multicolumn{1}{l|}{\textbf{Accuracy}} & \textbf{F1-Score} & \multicolumn{1}{l|}{\textbf{Accuracy}} & \textbf{F1-Score} \\ \hline
\multirow{4}{*}{25\%} & \multirow{2}{*}{\begin{tabular}[c]{@{}c@{}}4-way \\ 1-shot\end{tabular}} & Pre-train & \multicolumn{1}{l|}{0.2071} & 0.2035 & \multicolumn{1}{l|}{0.2148} & 0.2024 & \multicolumn{1}{l|}{0.2286} & 0.1900 & \multicolumn{1}{l|}{0.1911} & 0.1607 & \multicolumn{1}{l|}{\textbf{0.2881}} & \textbf{0.2716} & \multicolumn{1}{l|}{0.2233} & 0.1906 \\ \cline{3-15} 
 &  & re-train & \multicolumn{1}{l|}{0.2324} & 0.2290 & \multicolumn{1}{l|}{0.2338} & 0.2188 & \multicolumn{1}{l|}{0.2452} & 0.2444 & \multicolumn{1}{l|}{0.2462} & 0.2333 & \multicolumn{1}{l|}{\textbf{0.3019}} & \textbf{0.2867} & \multicolumn{1}{l|}{0.2100} & 0.2049 \\ \cline{2-15} 
 & \multirow{2}{*}{\begin{tabular}[c]{@{}c@{}}4-way \\ 5-shot\end{tabular}} & Pre-train & \multicolumn{1}{l|}{0.2690} & 0.2510 & \multicolumn{1}{l|}{0.2786} & 0.2480 & \multicolumn{1}{l|}{0.2100} & 0.2039 & \multicolumn{1}{l|}{0.2029} & 0.1861 & \multicolumn{1}{l|}{\textbf{0.3300}} & \textbf{0.3179} & \multicolumn{1}{l|}{0.2978} & 0.2705 \\ \cline{3-15} 
 &  & re-train & \multicolumn{1}{l|}{0.3129} & 0.3005 & \multicolumn{1}{l|}{\textbf{0.3824}} & \textbf{0.3761} & \multicolumn{1}{l|}{0.3290} & 0.2917 & \multicolumn{1}{l|}{0.2881} & 0.2467 & \multicolumn{1}{l|}{0.3681} & 0.3395 & \multicolumn{1}{l|}{0.3090} & 0.2892 \\ \hline
\multirow{4}{*}{50\%} & \multirow{2}{*}{\begin{tabular}[c]{@{}c@{}}4-way \\ 1-shot\end{tabular}} & Pre-train & \multicolumn{1}{l|}{0.3506} & 0.3461 & \multicolumn{1}{l|}{0.3456} & 0.3452 & \multicolumn{1}{l|}{0.3511} & 0.3354 & \multicolumn{1}{l|}{0.1914} & 0.1631 & \multicolumn{1}{l|}{\textbf{0.3844}} & \textbf{0.3506} & \multicolumn{1}{l|}{0.3119} & 0.2765 \\ \cline{3-15} 
 &  & re-train & \multicolumn{1}{l|}{0.3711} & 0.3605 & \multicolumn{1}{l|}{\textbf{0.4333}} & \textbf{0.4279} & \multicolumn{1}{l|}{0.3733} & 0.3598 & \multicolumn{1}{l|}{0.3111} & 0.3009 & \multicolumn{1}{l|}{0.3706} & 0.3614 & \multicolumn{1}{l|}{0.3678} & 0.3556 \\ \cline{2-15} 
 & \multirow{2}{*}{\begin{tabular}[c]{@{}c@{}}4-way \\ 5-shot\end{tabular}} & Pre-train & \multicolumn{1}{l|}{0.4117} & 0.3976 & \multicolumn{1}{l|}{0.3983} & 0.3989 & \multicolumn{1}{l|}{0.3617} & 0.3529 & \multicolumn{1}{l|}{0.2717} & 0.2609 & \multicolumn{1}{l|}{\textbf{0.4450}} & \textbf{0.4354} & \multicolumn{1}{l|}{0.3856} & 0.3580 \\ \cline{3-15} 
 &  & re-train & \multicolumn{1}{l|}{0.4311} & 0.4128 & \multicolumn{1}{l|}{0.5144} & 0.5130 & \multicolumn{1}{l|}{0.4661} & 0.4367 & \multicolumn{1}{l|}{0.4183} & 0.4058 & \multicolumn{1}{l|}{\textbf{0.5322}} & \textbf{0.5222 }& \multicolumn{1}{l|}{0.3944} & 0.3785 \\ \hline
\multirow{4}{*}{75\%} & \multirow{2}{*}{\begin{tabular}[c]{@{}c@{}}4-way \\ 1-shot\end{tabular}} & Pre-train & \multicolumn{1}{l|}{0.6233} & 0.6124 & \multicolumn{1}{l|}{0.5478} & 0.5270 & \multicolumn{1}{l|}{0.4267} & 0.4136 & \multicolumn{1}{l|}{0.4578} & 0.4557 & \multicolumn{1}{l|}{\textbf{0.7511}} & \textbf{0.7490} & \multicolumn{1}{l|}{0.5967} & 0.5786 \\ \cline{3-15} 
 &  & re-train & \multicolumn{1}{l|}{0.6589} & 0.6567 & \multicolumn{1}{l|}{0.6956} & 0.6962 & \multicolumn{1}{l|}{0.5911} & 0.5622 & \multicolumn{1}{l|}{0.6056} & 0.5837 & \multicolumn{1}{l|}{\textbf{0.8400}} & \textbf{0.8403 }& \multicolumn{1}{l|}{0.6833} & 0.6766 \\ \cline{2-15} 
 & \multirow{2}{*}{\begin{tabular}[c]{@{}c@{}}4-way \\ 5-shot\end{tabular}} & Pre-train & \multicolumn{1}{l|}{0.6878} & 0.6855 & \multicolumn{1}{l|}{0.6800} & 0.6797 & \multicolumn{1}{l|}{0.6122} & 0.6113 & \multicolumn{1}{l|}{0.5189} & 0.5192 & \multicolumn{1}{l|}{\textbf{0.7589}} & \textbf{0.7623} & \multicolumn{1}{l|}{0.6733} & 0.6584 \\ \cline{3-15} 
 &  & re-train & \multicolumn{1}{l|}{0.7078} & 0.7078 & \multicolumn{1}{l|}{0.7578} & 0.7566 & \multicolumn{1}{l|}{0.7211} & 0.7177 & \multicolumn{1}{l|}{0.6767} & 0.6716 & \multicolumn{1}{l|}{\textbf{0.8444}} & \textbf{0.8442} & \multicolumn{1}{l|}{0.7044} & 0.6933 \\ \hline
\end{tabular}}
\end{table}
%%%%%%%%%%%%%%%%%%%%%%%%%%%%%%%%%%%%%%%%%%%%%%%%%%%%%%%%%%%%%%%%%%

\normalsize
Similar to Table \ref{ATIS}, Table \ref{Web_Queries} illustrates performance of pre-trained and re-trained language models based prototype informed predictive pipelines across Web Queries dataset.
A deep analysis of Table \ref{Web_Queries}, reveals that among pre-trained language models based predictive pipelines, 3 language model based pedictive pipelines produced highest performance across all 3 data splits and both experimental settings. Models such as DistilBERT and DeBERTa-base showed limited effectiveness, with low precision and recall scores, reflecting their inability to generalize without task-specific tuning. Even Urdu-focused models like MuRIL and RoBERTa-small-Urdu struggled to achieve satisfactory results in their pre-trained state. However, re-training significantly improved performance by re-training the models on task-specific data, enhancing their accuracy and F1-Scores. This improvement was particularly evident for MuRIL and RoBERTa-small-Urdu, as their pre-training on Urdu-specific data allowed them to better adapt to the linguistic nuances of the target language.
The performance difference between 1-shot and 5-shot learning settings is evident, with models consistently achieving better results in 5-shot scenarios. For example, in the 4-way intent detection task with 25\% split, MuRIL’s accuracy improves from 0.2375 in the 1-shot setting to 0.2731 in the 5-shot setting. Similarly, BERT’s accuracy jumps from 0.2222 to 0.2911 in the same scenario. The primary advantage of the 5-shot approach is that it provides more examples per class, which helps the model learn more effectively by offering richer information. This is especially important for multilingual models like BERT and MuRIL, which need diverse examples to generalize well across different languages. On the other hand, the 1-shot method limits the models to a single example per intent, making it more difficult for them to generalize and perform optimally. For instance, in the 4-way task with 75\% split, MuRIL’s accuracy improves from 0.4142 in the 1-shot learning settings to 0.6303 in the 5-shot learning settings, while BERT improves from 0.5457 to 0.7073. These examples demonstrate how additional labeled examples enhance the models' ability to generalize and perform better, particularly in multilingual contexts.
Increasing the size of the training dataset leads to noticeable performance gains, particularly when moving from smaller datasets to larger ones. With only 25\% split, the performance of models is limited due to insufficient training examples. In this scenario, MuRIL's accuracy, for example, is 0.2375 in the 1-shot learning settings. Results improve moderately with 50\% split, where MuRIL’s accuracy increases to 0.2731 in the 5-shot setting. The most significant improvements are seen when using 75\% split, where models like MuRIL achieve their best performance. For instance, MuRIL's accuracy improves to 0.6303 in the 5-shot re-training scenario with 75\% split. This pattern shows how larger datasets, particularly with more examples per class, provide essential linguistic context, especially for low-resource languages like Urdu, helping models better generalize and adapt to task-specific nuances.

%%%%%%%%%%%%%%%%%%%%%%%%%%%%%%%%%%%%%%%%%%%%%%%%%%%%%%%%%%%%%%%%%%%%%%%

\normalsize
\subsection{Proposed framework predictive pipelines robustness analysis}\label{pre-recall-web-results}
\begin{figure*}[htbp]
\centering
\begin{tabular}{cc} 
\includegraphics[width=0.48\textwidth, trim={0 0 0.0cm 0}]
{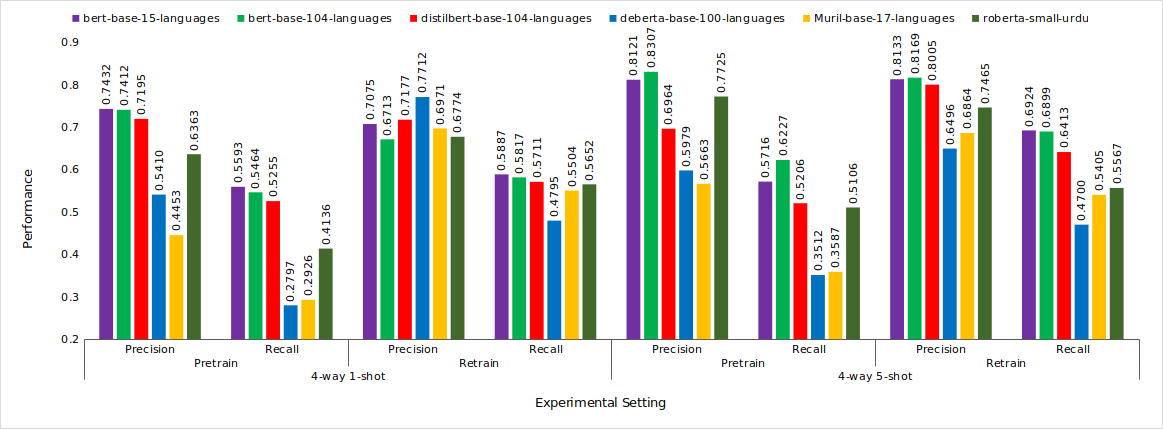} &
\includegraphics[width=0.48\textwidth, trim={0 0 0.0cm 0}]{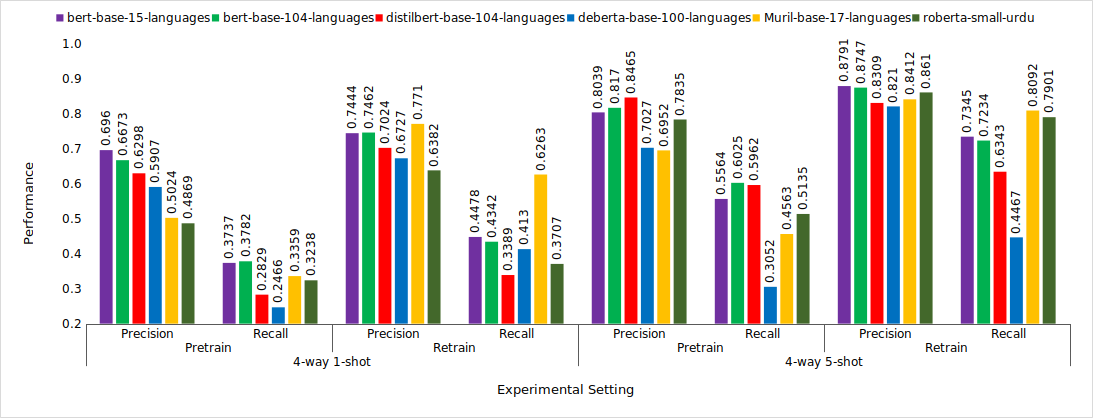} \\
{(a) 25\% Seen Classes}  & {(b) 50\% Seen Classes} \\[2pt]
\end{tabular}
\begin{tabular}{cc}
\includegraphics[width=0.48\textwidth, trim={0 0 0.0cm 0}]{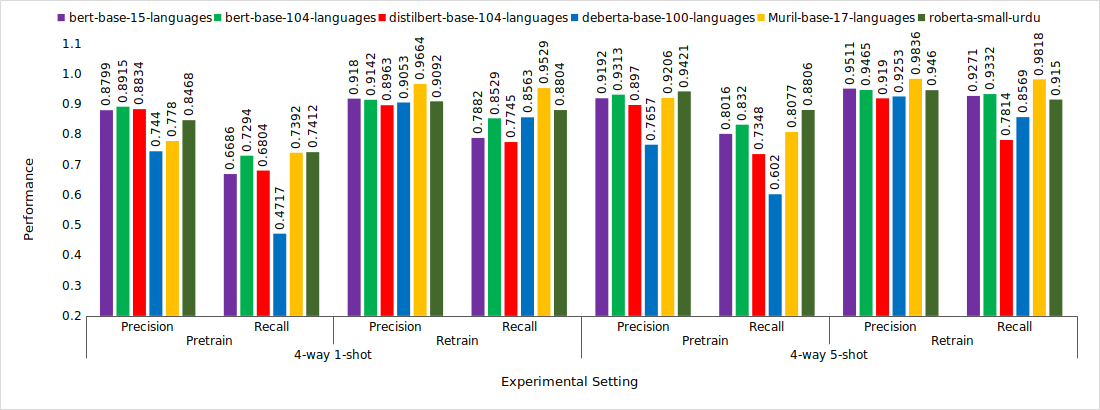} \\
{(c) 75\% Seen Classes}  \\ [2pt]
\end{tabular}
\caption{A Comprehensive Performance Analysis of Distinct Predictive Pipelines of Proposed Framework in terms of Precision and Recall over ATIS Dataset} 
\label{fig:atis_pre}
\end{figure*}

\begin{figure*}[htbp]
\centering
\begin{tabular}{cc} 

\includegraphics[width=0.48\textwidth, trim={0 0 0.0cm 0}]
{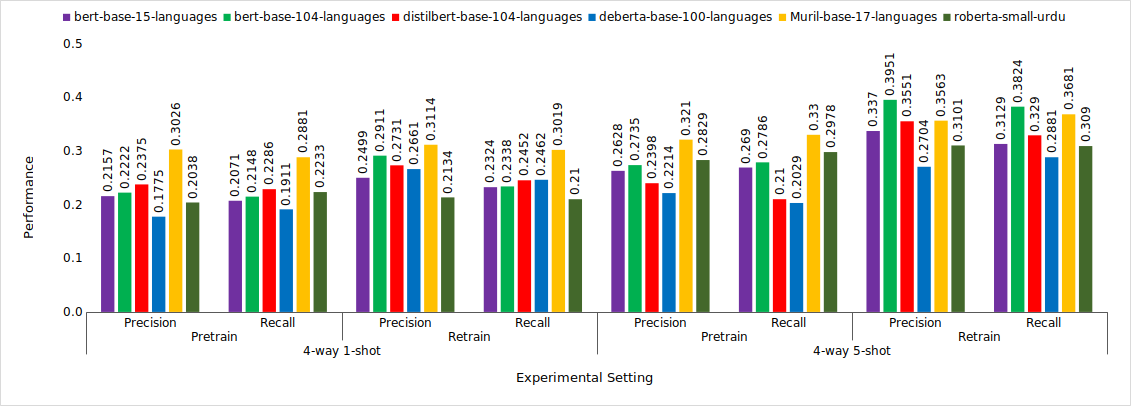} &
\includegraphics[width=0.48\textwidth, trim={0 0 0.0cm 0}]{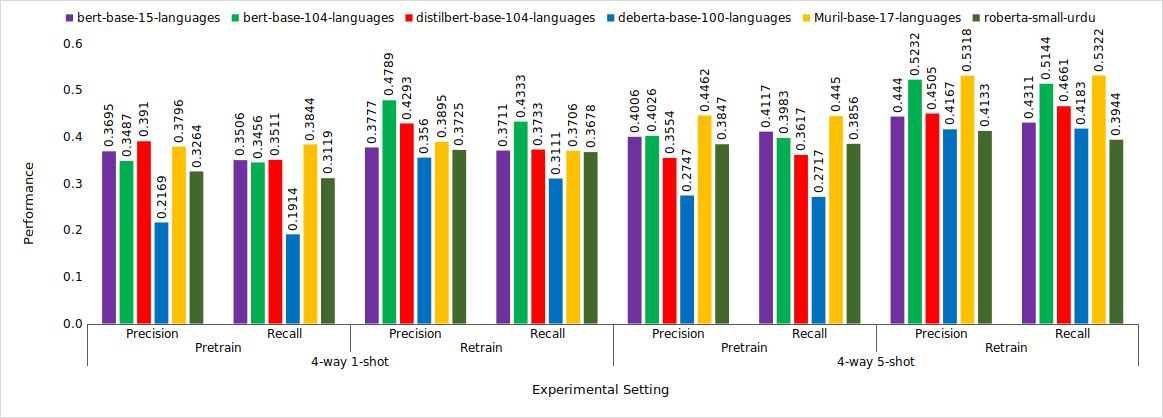} \\
{(a) 25\% Seen Classes}  & {(b) 50\% Seen Classes} \\[2pt]
\end{tabular}
\begin{tabular}{cc}
\includegraphics[width=0.48\textwidth, trim={0 0 0.0cm 0}]{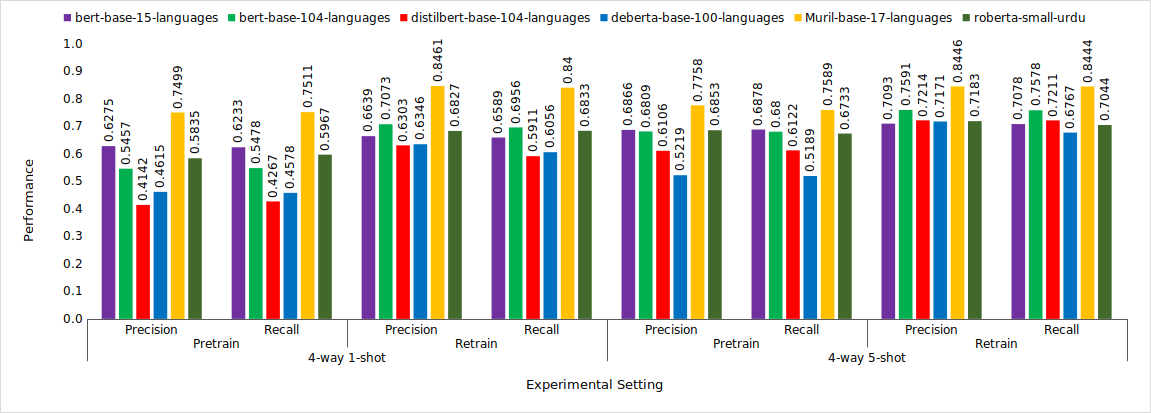} \\
{(c) 75\% Seen Classes}  \\ [2pt]
\end{tabular}
\caption{A Comprehensive Performance Analysis of Distinct Predictive Pipelines of Proposed Framework In terms of Precision and Recall over Web Queries dataset} 
\label{Web_Pre}
\end{figure*}

Following robustness criteria, framework predictive pipelines can be broadly classified into two categories, unbiased and biased. Biased predictors are those that exhibit either type 1 error or type 2 error. Type 1 Error (False Positive): Occurs when the predictor incorrectly predicts a positive class for an instance that actually belongs to the negative class, reflecting low precision. Type 2 Error (False Negative): Occurs when the predictor incorrectly predicts a negative class for an instance that belongs to the positive class, reflecting low recall. Biased predictors can be further categorized into three categories: High (point difference between precision and recall $\geq$ 5), Medium (point difference between precision and recall between 3 and 5), and Low (the point difference between precision and recall $<$ 3). A threshold of 0.01 was set for this categorization. It is evident from Figure \ref{Web_Pre} DeBERTa-100-languages exhibits a substantial bias, primarily due to its precision being significantly higher than its recall. This indicates a predominance of type II errors (false negatives), where the model fails to correctly identify certain intents despite their presence. In contrast, Muril-100-languages demonstrates balanced performance, with precision and recall being nearly identical. This reflects a negligible bias, as the model maintains a low rate of both type I errors (false positives) and type II errors (false negatives). As for BERT-base-104-languages, it is moderately biased, with a noticeable imbalance between precision and recall, similar to DistilBERT-104-languages, but with slightly less discrepancy in performance. Roberta-small-Urdu also shows moderate imbalance, especially during pre-training.\\
It is evident from Figure \ref{fig:atis_pre} that similar to Web Queries dataset, DeBERTa-100-languages exhibits a high bias for ATIS dataset as well, as recall is consistently lower than precision across different configurations. This indicates a prevalence of Type II errors (false negatives), where the model fails to correctly identify certain positive instances. This bias is especially evident in low-shot scenario, such as 4-way 1-shot, where the recall values are significantly lower. On the other hand, Muril-100-languages demonstrates balanced performance, as precision and recall values are nearly identical across configurations. This balance indicates a low rate of both Type I errors (false positives) and Type II errors (false negatives), making it one of the most reliable models in terms of bias. BERT-base-104-languages and DistilBERT-104-languages show moderate bias, with noticeable differences between precision and recall. In these cases, recall is lower than precision, suggesting a tendency toward Type II errors. However, the degree of imbalance is not as severe as in DeBERTa-100-languages. Similarly, RoBERTa-small-Urdu also displays moderate bias, particularly during pre-training stage, where precision slightly exceeds recall, leading to a mix of Type I and Type II errors, though Type II errors are more pronounced. Finally, BERT-base-15-languages shows consistently low bias, with precision and recall values closely aligned across all configurations. This indicates a minimal occurrence of both error types, reflecting a strong and balanced predictive capability, especially as the data split increases. Overall, while Muril-100-languages and BERT-base-15-languages stand out as balanced performers, DeBERTa-100-languages struggles with high bias and frequent Type II errors.
%%%%%%%%%%%%%%%%%%%%%%%%%%%%%%%%%%%%%%%%%%%%%%%%%%%%%%%%%%%%%

%%%%%%%%%%%%%%%%%%%%%%%%%%%%%%%%%%%%%%%%%%%%%%%%%%%%%%%%%
%%%%%%%%%%%%%%%%%%%%%%%%%%%%%%%%%%%

%%%%%%%%%%%%%%%%%%%%%%%%%%%%%%%%%%%%%%%%%%%%%%%%%%%%%%%%%

\subsection{Performance Impact of different Similarity Metrics on Distinct Predictive Pipelines of Proposed framework}
The choice of similarity measure is critical to determine the effectiveness of the proposed framework, as it quantifies the relationship between the embeddings of support and query samples. To analyze impact of similarity metric on the performance of the different predictive pipelines we experimented with 13 different similarity metrics. These similarity metrics include: angular similarity \cite{apaydin2006access}, bhattacharyya distance \cite{kazakos1978bhattacharyya}, chebyshev distance \cite{coghetto2016chebyshev}, cosine similarity \cite{rahutomo2012semantic}, dice similarity \cite{ye2014dice}, dot product \cite{smola2000regularization}, euclidean distance \cite{danielsson1980euclidean}, hamming distance \cite{bookstein2002generalized}, jaccard similarity \cite{ivchenko1998jaccard}, KL divergence \cite{van2014renyi}, L2 distance \cite{ruschendorf1990characterization}, manhattan distance \cite{kobayashi2014similarity} and pearson correlation \cite{cohen2009pearson}. Table \ref{ATIS_Sim} illustrates F1-Score of 13 distinct similarity metric across 6 distinct predictive pipelines (re-trained models) under 4-way 1-shot and 4-way 5-shot experimental settings over different data splits for ATIS dataset. Supplementary file (ATIS\_Dataset) presents performance of different similarity metrics across 4 different measures.

It is evident from Table \ref{ATIS_Sim} performance of similarity metric vary significantly in 4-way 1-shot setting. Specifically, Chebyshev distance, jaccard similarity and hamming distance underperforms especially with 25\% data. These methods do not seem to provide strong discriminative power with only one example per class, specifically for bert-base-15-languages model. Similarly, dot product similarity and KL divergence also exhibit consistently low values across different models, exhibit poor performance. As these measures do not appear to be effective in capturing meaningful distinctions in low-data scenarios. However, cosine similarity and angular similarity stand out as the top performers across most configurations. These two measures maintain a relatively steady level of performance even with limited data (only one example per class) and remains competitive throughout the dataset splits. This indicates that these similarity measures are relatively resilient in low-data settings which makes them suitable for one-shot learning tasks.
Performance of most similarity measures improves significantly in 4-way 5-shot setting, as the models have five examples per class as labelled data. Cosine similarity continues to exhibit strong performance across all configurations, especially for models like bert-base-104-languages and bert-base-15-languages. Angular similarity also performs consistently well in 4-way 5-shot setting, showing a noticeable improvement over 4-way 1-shot performance. However, while it is competitive, it does not surpass cosine similarity in mostly configurations. However, L2 distance and dot product similarity show more improvement in 4-way 5-shot setting compared to 4-way 1-shot, with dot product similarity reaching 0.7384 for bert-base-104-languages at 50\% data split, and L2 distance demostarte an improvement in models like bert-base-104-languages at 75\%. Still, these measures don't consistently match the performance of cosine similarity. Some similarity measures, like chebyshev distance, bhattacharyya distance, and manhattan similarity, continue to show less reliability.They are still inconsistent and do not show the same level of improvement as others. 
In a nutshell, while most measures show improvement in the 4-way 5-shot setting, cosine similarity remains the dominant performer, showing the most robust and consistent results. This is primarily due to its ability to measure the directional alignment between vectors, rather than focusing on their absolute magnitude. In high-dimensional spaces of NLP tasks, the length of the vectors can vary substantially. However, cosine similarity remains unaffected by these variations that make it more stable and interpretable.

%%%%%%%%%%%%%%%%%%%%%%%%%%%%%%%%%%%%%%%%%%%%%%%%%%%%%%%%%%%%%%%%%%%%%%%%%%%%%%%

\begin{table}[htbp]
\caption{F1-Score of 4-way 1-shot and 4-way 5-shot over ATIS dataset}
\label{ATIS_Sim}
\renewcommand{\arraystretch}{1.5}
\resizebox{1.0\textwidth}{!}{
\begin{tabular}{|l|c|cccccc|cccccc|}
\hline
\multicolumn{2}{|c|}{\textbf{Experimental Setting}} & \multicolumn{6}{c|}{\textbf{4-way 1-shot}} & \multicolumn{6}{c|}{\textbf{4-way 5-shot}} \\ \hline
\textbf{Model} & {\textbf{Seen Classes}} & \multicolumn{1}{l|}{\textbf{\begin{tabular}[c]{@{}l@{}}BERT-\\ base-\\ 15-\\ languages\end{tabular}}} & \multicolumn{1}{c|}{\textbf{\begin{tabular}[c]{@{}l@{}}BERT-\\ base-\\ 104-\\ languages\end{tabular}}} & \multicolumn{1}{c|}{\textbf{\begin{tabular}[c]{@{}l@{}}DistilBERT-\\ base-\\ 104-\\ languages\end{tabular}}} & \multicolumn{1}{c|}{\textbf{\begin{tabular}[c]{@{}l@{}}DeBERTa-\\ base-\\ 100-\\ languages\end{tabular}}} & \multicolumn{1}{c|}{\textbf{\begin{tabular}[c]{@{}l@{}}MURiL-\\ base-\\ 100-\\ languages\end{tabular}}} & \textbf{\begin{tabular}[c]{@{}l@{}}RoBERTa-\\ small-\\ Urdu\end{tabular}} & \multicolumn{1}{c|}{\textbf{\begin{tabular}[c]{@{}l@{}}BERT-\\ base-\\ 15-\\ languages\end{tabular}}} & \multicolumn{1}{c|}{\textbf{\begin{tabular}[c]{@{}l@{}}BERT-\\ base-\\ 104-\\ languages\end{tabular}}} & \multicolumn{1}{c|}{\textbf{\begin{tabular}[c]{@{}l@{}}DistilBERT-\\ base-\\ 104-\\ languages\end{tabular}}} & \multicolumn{1}{c|}{\textbf{\begin{tabular}[c]{@{}l@{}}DeBERTa-\\ base-\\ 100-\\ languages\end{tabular}}} & \multicolumn{1}{c|}{\textbf{\begin{tabular}[c]{@{}l@{}}MURiL-\\ base-\\ 100-\\ languages\end{tabular}}} & \textbf{\begin{tabular}[c]{@{}l@{}}RoBERTa-\\ small-\\ Urdu\end{tabular}} \\ \hline
\multirow{3}{*}{\begin{tabular}[c]{@{}l@{}}Angular\\ Similarity\end{tabular}} & 25 & \multicolumn{1}{l|}{0.4833} & \multicolumn{1}{l|}{0.4220} & \multicolumn{1}{l|}{0.3996} & \multicolumn{1}{l|}{0.4217} & \multicolumn{1}{l|}{0.4881} & 0.2217 & \multicolumn{1}{l|}{0.4234} & \multicolumn{1}{l|}{0.4829} & \multicolumn{1}{l|}{0.3936} & \multicolumn{1}{l|}{0.3502} & \multicolumn{1}{l|}{0.3971} & 0.4690 \\ \cline{2-14} 
 & 50 & \multicolumn{1}{l|}{0.4556} & \multicolumn{1}{l|}{0.3837} & \multicolumn{1}{l|}{0.2566} & \multicolumn{1}{l|}{0.1546} & \multicolumn{1}{l|}{0.2803} & 0.3561 & \multicolumn{1}{l|}{0.6109} & \multicolumn{1}{l|}{0.7485} & \multicolumn{1}{l|}{0.5951} & \multicolumn{1}{l|}{0.4574} & \multicolumn{1}{l|}{0.6753} & 0.7653 \\ \cline{2-14} 
 & 75 & \multicolumn{1}{l|}{0.7246} & \multicolumn{1}{l|}{0.8020} & \multicolumn{1}{l|}{0.7032} & \multicolumn{1}{l|}{0.8018} & \multicolumn{1}{l|}{0.9061} & 0.8013 & \multicolumn{1}{l|}{0.8150} & \multicolumn{1}{l|}{0.9514} & \multicolumn{1}{l|}{0.8316} & \multicolumn{1}{l|}{0.8198} & \multicolumn{1}{l|}{0.9774} & 0.9171 \\ \hline
\multirow{3}{*}{\begin{tabular}[c]{@{}l@{}}Bhattacharyya\\ Distance\end{tabular}} & 25 & \multicolumn{1}{l|}{0.0180} & \multicolumn{1}{l|}{0.0208} & \multicolumn{1}{l|}{0.0128} & \multicolumn{1}{l|}{0.0759} & \multicolumn{1}{l|}{0.1697} & 0.0853 & \multicolumn{1}{l|}{0.0257} & \multicolumn{1}{l|}{0.0256} & \multicolumn{1}{l|}{0.0196} & \multicolumn{1}{l|}{0.1352} & \multicolumn{1}{l|}{0.2187} & 0.1526 \\ \cline{2-14} 
 & 50 & \multicolumn{1}{l|}{0.0939} & \multicolumn{1}{l|}{0.1290} & \multicolumn{1}{l|}{0.0201} & \multicolumn{1}{l|}{0.2092} & \multicolumn{1}{l|}{0.0464} & 0.1840 & \multicolumn{1}{l|}{0.1410} & \multicolumn{1}{l|}{0.1493} & \multicolumn{1}{l|}{0.1011} & \multicolumn{1}{l|}{0.1295} & \multicolumn{1}{l|}{0.0700} & 0.1720 \\ \cline{2-14} 
 & 75 & \multicolumn{1}{l|}{0.4730} & \multicolumn{1}{l|}{0.1682} & \multicolumn{1}{l|}{0.1653} & \multicolumn{1}{l|}{0.2111} & \multicolumn{1}{l|}{0.6689} & 0.3442 & \multicolumn{1}{l|}{0.5526} & \multicolumn{1}{l|}{0.2117} & \multicolumn{1}{l|}{0.6918} & \multicolumn{1}{l|}{0.3186} & \multicolumn{1}{l|}{0.5415} & 0.3512 \\ \hline
\multirow{3}{*}{\begin{tabular}[c]{@{}l@{}}Chebyshev\\ Distance\end{tabular}} & 25 & \multicolumn{1}{l|}{0.0613} & \multicolumn{1}{l|}{0.0026} & \multicolumn{1}{l|}{0.0010} & \multicolumn{1}{l|}{0.0298} & \multicolumn{1}{l|}{0.0034} & 0.1732 & \multicolumn{1}{l|}{0.0013} & \multicolumn{1}{l|}{0.0772} & \multicolumn{1}{l|}{0.1871} & \multicolumn{1}{l|}{0.0002} & \multicolumn{1}{l|}{0.1871} & 0.0103 \\ \cline{2-14} 
 & 50 & \multicolumn{1}{l|}{0.0123} & \multicolumn{1}{l|}{0.0042} & \multicolumn{1}{l|}{0.0016} & \multicolumn{1}{l|}{0.0119} & \multicolumn{1}{l|}{0.0119} & 0.0042 & \multicolumn{1}{l|}{0.0002} & \multicolumn{1}{l|}{0.0002} & \multicolumn{1}{l|}{0.0022} & \multicolumn{1}{l|}{0.0050} & \multicolumn{1}{l|}{0.0022} & 0.0011 \\ \cline{2-14} 
 & 75 & \multicolumn{1}{l|}{0.5735} & \multicolumn{1}{l|}{0.1806} & \multicolumn{1}{l|}{0.0070} & \multicolumn{1}{l|}{0.0070} & \multicolumn{1}{l|}{0.4123} & 0.0009 & \multicolumn{1}{l|}{0.4246} & \multicolumn{1}{l|}{0.0004} & \multicolumn{1}{l|}{0.4246} & \multicolumn{1}{l|}{0.0043} & \multicolumn{1}{l|}{0.1963} & 0.0196 \\ \hline
\multirow{3}{*}{\begin{tabular}[c]{@{}l@{}}Cosine\\ Similarity\end{tabular}} & 25 & \multicolumn{1}{l|}{\textbf{0.6181}} & \multicolumn{1}{l|}{\textbf{0.6088}} & \multicolumn{1}{l|}{\textbf{0.6083}} & \multicolumn{1}{l|}{0.4911} & \multicolumn{1}{l|}{0.5661} & \textbf{0.5920} & \multicolumn{1}{l|}{0.7222} & \multicolumn{1}{l|}{0.7084} & \multicolumn{1}{l|}{\textbf{0.6711}} & \multicolumn{1}{l|}{0.5018} & \multicolumn{1}{l|}{\textbf{0.5826}} & 0.5982 \\ \cline{2-14} 
 & 50 & \multicolumn{1}{l|}{\textbf{0.4708}} & \multicolumn{1}{l|}{0.4975} & \multicolumn{1}{l|}{0.3506} & \multicolumn{1}{l|}{0.4652} & \multicolumn{1}{l|}{0.6652} & 0.3600 & \multicolumn{1}{l|}{\textbf{0.7815}} & \multicolumn{1}{l|}{0.7693} & \multicolumn{1}{l|}{0.6502} & \multicolumn{1}{l|}{0.5004} & \multicolumn{1}{l|}{0.8175} & 0.8161 \\ \cline{2-14} 
 & 75 & \multicolumn{1}{l|}{0.8234} & \multicolumn{1}{l|}{0.8675} & \multicolumn{1}{l|}{\textbf{0.8100}} & \multicolumn{1}{l|}{0.8638} & \multicolumn{1}{l|}{0.9581} & \textbf{0.8879} & \multicolumn{1}{l|}{0.9344} & \multicolumn{1}{l|}{0.9371} & \multicolumn{1}{l|}{0.8114} & \multicolumn{1}{l|}{0.8757} & \multicolumn{1}{l|}{\textbf{0.9825}} & 0.9251 \\ \hline
\multirow{3}{*}{\begin{tabular}[c]{@{}l@{}}Dice\\ Similarity\end{tabular}} & 25 & \multicolumn{1}{l|}{0.0621} & \multicolumn{1}{l|}{0.1896} & \multicolumn{1}{l|}{0.0561} & \multicolumn{1}{l|}{0.0627} & \multicolumn{1}{l|}{0.1319} & 0.0540 & \multicolumn{1}{l|}{0.2876} & \multicolumn{1}{l|}{0.2757} & \multicolumn{1}{l|}{0.1583} & \multicolumn{1}{l|}{0.1548} & \multicolumn{1}{l|}{0.1540} & 0.1087 \\ \cline{2-14} 
 & 50 & \multicolumn{1}{l|}{0.2636} & \multicolumn{1}{l|}{0.2670} & \multicolumn{1}{l|}{0.3753} & \multicolumn{1}{l|}{0.1176} & \multicolumn{1}{l|}{0.5250} & 0.2302 & \multicolumn{1}{l|}{0.3985} & \multicolumn{1}{l|}{0.5259} & \multicolumn{1}{l|}{0.3682} & \multicolumn{1}{l|}{0.3780} & \multicolumn{1}{l|}{0.6180} & 0.4721 \\ \cline{2-14} 
 & 75 & \multicolumn{1}{l|}{0.6530} & \multicolumn{1}{l|}{0.8439} & \multicolumn{1}{l|}{0.7004} & \multicolumn{1}{l|}{0.7369} & \multicolumn{1}{l|}{0.6652} & 0.8081 & \multicolumn{1}{l|}{0.7134} & \multicolumn{1}{l|}{0.8582} & \multicolumn{1}{l|}{0.6355} & \multicolumn{1}{l|}{0.7746} & \multicolumn{1}{l|}{0.8143} & 0.8040 \\ \hline
\multirow{3}{*}{\begin{tabular}[c]{@{}l@{}}Dot product\\ Similarity\end{tabular}} & 25 & \multicolumn{1}{l|}{0.4649} & \multicolumn{1}{l|}{0.5537} & \multicolumn{1}{l|}{0.4593} & \multicolumn{1}{l|}{0.4916} & \multicolumn{1}{l|}{0.3906} & 0.4722 & \multicolumn{1}{l|}{0.6880} & \multicolumn{1}{l|}{0.6283} & \multicolumn{1}{l|}{0.6033} & \multicolumn{1}{l|}{0.4874} & \multicolumn{1}{l|}{0.4507} & 0.6491 \\ \cline{2-14} 
 & 50 & \multicolumn{1}{l|}{0.3880} & \multicolumn{1}{l|}{0.5381} & \multicolumn{1}{l|}{0.3503} & \multicolumn{1}{l|}{0.5552} & \multicolumn{1}{l|}{0.4088} & 0.2803 & \multicolumn{1}{l|}{0.7384} & \multicolumn{1}{l|}{0.7224} & \multicolumn{1}{l|}{\textbf{0.8353}} & \multicolumn{1}{l|}{\textbf{0.7190}} & \multicolumn{1}{l|}{0.7720} & 0.8152 \\ \cline{2-14} 
 & 75 & \multicolumn{1}{l|}{\textbf{0.8374}} & \multicolumn{1}{l|}{\textbf{0.9419}} & \multicolumn{1}{l|}{0.7264} & \multicolumn{1}{l|}{0.8240} & \multicolumn{1}{l|}{\textbf{0.9636}} & 0.7289 & \multicolumn{1}{l|}{0.9426} & \multicolumn{1}{l|}{0.9478} & \multicolumn{1}{l|}{0.9293} & \multicolumn{1}{l|}{0.9262} & \multicolumn{1}{l|}{0.9672} & 0.9374 \\ \hline
\multirow{3}{*}{\begin{tabular}[c]{@{}l@{}}Euclidean\\ Distance\end{tabular}} & 25 & \multicolumn{1}{l|}{0.6073} & \multicolumn{1}{l|}{0.5451} & \multicolumn{1}{l|}{0.5156} & \multicolumn{1}{l|}{\textbf{0.5049}} & \multicolumn{1}{l|}{0.3377} & 0.4403 & \multicolumn{1}{l|}{0.6320} & \multicolumn{1}{l|}{0.6286} & \multicolumn{1}{l|}{0.5261} & \multicolumn{1}{l|}{0.4759} & \multicolumn{1}{l|}{0.4521} & 0.6980 \\ \cline{2-14} 
 & 50 & \multicolumn{1}{l|}{0.4252} & \multicolumn{1}{l|}{0.4715} & \multicolumn{1}{l|}{\textbf{0.4606}} & \multicolumn{1}{l|}{0.3269} & \multicolumn{1}{l|}{0.3868} & 0.4337 & \multicolumn{1}{l|}{0.7453} & \multicolumn{1}{l|}{0.8154} & \multicolumn{1}{l|}{0.7104} & \multicolumn{1}{l|}{0.4889} & \multicolumn{1}{l|}{0.7430} & 0.8170 \\ \cline{2-14} 
 & 75 & \multicolumn{1}{l|}{0.8057} & \multicolumn{1}{l|}{0.8747} & \multicolumn{1}{l|}{0.7025} & \multicolumn{1}{l|}{0.8833} & \multicolumn{1}{l|}{0.8922} & 0.7325 & \multicolumn{1}{l|}{0.9504} & \multicolumn{1}{l|}{0.9238} & \multicolumn{1}{l|}{0.8875} & \multicolumn{1}{l|}{0.7564} & \multicolumn{1}{l|}{0.9604} & \textbf{0.9393} \\ \hline
\multirow{3}{*}{\begin{tabular}[c]{@{}l@{}}Hamming\\ Distance\end{tabular}} & 25 & \multicolumn{1}{l|}{0.0003} & \multicolumn{1}{l|}{0.0003} & \multicolumn{1}{l|}{0.0003} & \multicolumn{1}{l|}{0.0003} & \multicolumn{1}{l|}{0.0003} & 0.0003 & \multicolumn{1}{l|}{0.0002} & \multicolumn{1}{l|}{0.0002} & \multicolumn{1}{l|}{0.0002} & \multicolumn{1}{l|}{0.0002} & \multicolumn{1}{l|}{0.0002} & 0.0002 \\ \cline{2-14} 
 & 50 & \multicolumn{1}{l|}{0.0005} & \multicolumn{1}{l|}{0.0005} & \multicolumn{1}{l|}{0.0005} & \multicolumn{1}{l|}{0.0005} & \multicolumn{1}{l|}{0.0005} & 0.0005 & \multicolumn{1}{l|}{0.0002} & \multicolumn{1}{l|}{0.0002} & \multicolumn{1}{l|}{0.0002} & \multicolumn{1}{l|}{0.0002} & \multicolumn{1}{l|}{0.0002} & 0.0002 \\ \cline{2-14} 
 & 75 & \multicolumn{1}{l|}{0.0009} & \multicolumn{1}{l|}{0.0009} & \multicolumn{1}{l|}{0.0009} & \multicolumn{1}{l|}{0.0009} & \multicolumn{1}{l|}{0.0009} & 0.0009 & \multicolumn{1}{l|}{0.0004} & \multicolumn{1}{l|}{0.0004} & \multicolumn{1}{l|}{0.0004} & \multicolumn{1}{l|}{0.0004} & \multicolumn{1}{l|}{0.0004} & 0.0004 \\ \hline
\multirow{3}{*}{\begin{tabular}[c]{@{}l@{}}Jaccard\\ Similarity\end{tabular}} & 25 & \multicolumn{1}{l|}{0.1258} & \multicolumn{1}{l|}{0.2312} & \multicolumn{1}{l|}{0.1543} & \multicolumn{1}{l|}{0.0591} & \multicolumn{1}{l|}{0.1120} & 0.1854 & \multicolumn{1}{l|}{0.2135} & \multicolumn{1}{l|}{0.2926} & \multicolumn{1}{l|}{0.1150} & \multicolumn{1}{l|}{0.0903} & \multicolumn{1}{l|}{0.1655} & 0.0885 \\ \cline{2-14} 
 & 50 & \multicolumn{1}{l|}{0.3677} & \multicolumn{1}{l|}{0.3421} & \multicolumn{1}{l|}{0.1968} & \multicolumn{1}{l|}{\textbf{0.5680}} & \multicolumn{1}{l|}{0.3937} & \textbf{0.4633} & \multicolumn{1}{l|}{0.4344} & \multicolumn{1}{l|}{0.5020} & \multicolumn{1}{l|}{0.2601} & \multicolumn{1}{l|}{0.4716} & \multicolumn{1}{l|}{0.5349} & 0.3562 \\ \cline{2-14} 
 & 75 & \multicolumn{1}{l|}{0.7449} & \multicolumn{1}{l|}{0.7397} & \multicolumn{1}{l|}{0.6554} & \multicolumn{1}{l|}{0.7897} & \multicolumn{1}{l|}{0.6571} & 0.6836 & \multicolumn{1}{l|}{0.5111} & \multicolumn{1}{l|}{0.6109} & \multicolumn{1}{l|}{0.6909} & \multicolumn{1}{l|}{0.7929} & \multicolumn{1}{l|}{0.3684} & 0.5902 \\ \hline
\multirow{3}{*}{\begin{tabular}[c]{@{}l@{}}KL\\ Divergence\end{tabular}} & 25 & \multicolumn{1}{l|}{0.0003} & \multicolumn{1}{l|}{0.0003} & \multicolumn{1}{l|}{0.0003} & \multicolumn{1}{l|}{0.0003} & \multicolumn{1}{l|}{0.0003} & 0.0003 & \multicolumn{1}{l|}{0.0002} & \multicolumn{1}{l|}{0.0002} & \multicolumn{1}{l|}{0.0002} & \multicolumn{1}{l|}{0.0002} & \multicolumn{1}{l|}{0.0002} & 0.0002 \\ \cline{2-14} 
 & 50 & \multicolumn{1}{l|}{0.0005} & \multicolumn{1}{l|}{0.0005} & \multicolumn{1}{l|}{0.0005} & \multicolumn{1}{l|}{0.0005} & \multicolumn{1}{l|}{0.0005} & 0.0005 & \multicolumn{1}{l|}{0.0002} & \multicolumn{1}{l|}{0.0002} & \multicolumn{1}{l|}{0.0002} & \multicolumn{1}{l|}{0.0002} & \multicolumn{1}{l|}{0.0002} & 0.0002 \\ \cline{2-14} 
 & 75 & \multicolumn{1}{l|}{0.0009} & \multicolumn{1}{l|}{0.0009} & \multicolumn{1}{l|}{0.0009} & \multicolumn{1}{l|}{0.0009} & \multicolumn{1}{l|}{0.0009} & 0.0009 & \multicolumn{1}{l|}{0.0004} & \multicolumn{1}{l|}{0.0004} & \multicolumn{1}{l|}{0.0004} & \multicolumn{1}{l|}{0.0004} & \multicolumn{1}{l|}{0.0004} & 0.0004 \\ \hline
\multirow{3}{*}{\begin{tabular}[c]{@{}l@{}}L2\\ Distance\end{tabular}} & 25 & \multicolumn{1}{l|}{0.4210} & \multicolumn{1}{l|}{0.5224} & \multicolumn{1}{l|}{0.4136} & \multicolumn{1}{l|}{0.4180} & \multicolumn{1}{l|}{0.4825} & 0.5353 & \multicolumn{1}{l|}{0.6041} & \multicolumn{1}{l|}{0.5637} & \multicolumn{1}{l|}{0.6251} & \multicolumn{1}{l|}{0.5246} & \multicolumn{1}{l|}{0.4526} & \textbf{0.7224} \\ \cline{2-14} 
 & 50 & \multicolumn{1}{l|}{0.3485} & \multicolumn{1}{l|}{0.4111} & \multicolumn{1}{l|}{0.3272} & \multicolumn{1}{l|}{0.3902} & \multicolumn{1}{l|}{0.5005} & 0.4047 & \multicolumn{1}{l|}{0.6673} & \multicolumn{1}{l|}{0.7427} & \multicolumn{1}{l|}{0.6117} & \multicolumn{1}{l|}{0.5986} & \multicolumn{1}{l|}{0.7046} & \textbf{0.8271} \\ \cline{2-14} 
 & 75 & \multicolumn{1}{l|}{0.7968} & \multicolumn{1}{l|}{0.8954} & \multicolumn{1}{l|}{0.7941} & \multicolumn{1}{l|}{0.7759} & \multicolumn{1}{l|}{0.9608} & 0.7362 & \multicolumn{1}{l|}{\textbf{0.9684}} & \multicolumn{1}{l|}{0.9317} & \multicolumn{1}{l|}{\textbf{0.9428}} & \multicolumn{1}{l|}{0.8590} & \multicolumn{1}{l|}{0.9721} & 0.9344 \\ \hline
\multirow{3}{*}{\begin{tabular}[c]{@{}l@{}}Manhattan\\ Distance\end{tabular}} & 25 & \multicolumn{1}{l|}{0.3954} & \multicolumn{1}{l|}{0.4742} & \multicolumn{1}{l|}{0.2529} & \multicolumn{1}{l|}{0.4659} & \multicolumn{1}{l|}{0.1058} & 0.4396 & \multicolumn{1}{l|}{0.6338} & \multicolumn{1}{l|}{0.4394} & \multicolumn{1}{l|}{0.3039} & \multicolumn{1}{l|}{0.3963} & \multicolumn{1}{l|}{0.2506} & 0.3018 \\ \cline{2-14} 
 & 50 & \multicolumn{1}{l|}{0.4219} & \multicolumn{1}{l|}{\textbf{0.5284}} & \multicolumn{1}{l|}{0.3947} & \multicolumn{1}{l|}{0.4812} & \multicolumn{1}{l|}{0.6434} & 0.4015 & \multicolumn{1}{l|}{0.7019} & \multicolumn{1}{l|}{\textbf{0.8187}} & \multicolumn{1}{l|}{0.6737} & \multicolumn{1}{l|}{0.4701} & \multicolumn{1}{l|}{0.5119} & 0.7622 \\ \cline{2-14} 
 & 75 & \multicolumn{1}{l|}{0.8326} & \multicolumn{1}{l|}{0.9084} & \multicolumn{1}{l|}{0.8084} & \multicolumn{1}{l|}{\textbf{0.9036}} & \multicolumn{1}{l|}{0.8994} & 0.8286 & \multicolumn{1}{l|}{0.9504} & \multicolumn{1}{l|}{\textbf{0.9547}} & \multicolumn{1}{l|}{0.9314} & \multicolumn{1}{l|}{\textbf{0.9346}} & \multicolumn{1}{l|}{0.6905} & 0.9073 \\ \hline
\multirow{3}{*}{\begin{tabular}[c]{@{}l@{}}Pearson\\ Correlation\end{tabular}} & 25 & \multicolumn{1}{l|}{0.6091} & \multicolumn{1}{l|}{0.5910} & \multicolumn{1}{l|}{0.5750} & \multicolumn{1}{l|}{0.4983} & \multicolumn{1}{l|}{\textbf{0.5823}} & 0.5862 & \multicolumn{1}{l|}{\textbf{0.7264}} & \multicolumn{1}{l|}{\textbf{0.7099}} & \multicolumn{1}{l|}{0.5491} & \multicolumn{1}{l|}{\textbf{0.5250}} & \multicolumn{1}{l|}{0.5639} & 0.6372 \\ \cline{2-14} 
 & 50 & \multicolumn{1}{l|}{0.4576} & \multicolumn{1}{l|}{0.4904} & \multicolumn{1}{l|}{0.3496} & \multicolumn{1}{l|}{0.4594} & \multicolumn{1}{l|}{\textbf{0.6660}} & 0.3548 & \multicolumn{1}{l|}{0.7768} & \multicolumn{1}{l|}{0.7691} & \multicolumn{1}{l|}{0.6879} & \multicolumn{1}{l|}{0.4990} & \multicolumn{1}{l|}{\textbf{0.8201}} & 0.8137 \\ \cline{2-14} 
 & 75 & \multicolumn{1}{l|}{0.8228} & \multicolumn{1}{l|}{0.8567} & \multicolumn{1}{l|}{\textbf{0.8100}} & \multicolumn{1}{l|}{0.8757} & \multicolumn{1}{l|}{0.9601} & 0.8735 & \multicolumn{1}{l|}{0.9358} & \multicolumn{1}{l|}{0.9371} & \multicolumn{1}{l|}{0.8139} & \multicolumn{1}{l|}{0.8638} & \multicolumn{1}{l|}{\textbf{0.9825}} & 0.9248 \\ \hline
\end{tabular}}
\end{table}

%%%%%%%%%%%%%%%%%%%%%%%%%%%%%%%%%%%%%%%%%%%%%%%%%%%%%%%%%%%%%%%%%%%%%%%%%%%%%%
\normalsize

Similar to Table \ref{ATIS_Sim}, Table \ref{Web_Sim} presents F1-Score of 6 distinct re-trained language models across 13 various similarity measure under 4-way 1-shot and 4-way 5-shot settings across 3 different data splits (25\%, 50\% and 75\% seen data) for Web Queries dataset. Additionally, Supplementary file  (Web\_Dataset) illustrates a detailed performance of 13 distinct similarity measures across 4 different evaluation measures.
A deep analysis of Table \ref{Web_Sim} reveals that, in the 4-way 1-shot setting, performance varies depending on the chosen similarity measure. It is evident that bhattacharyya distance \cite{kazakos1978bhattacharyya}, chebyshev distance \cite{coghetto2016chebyshev}, hamming distance \cite{bookstein2002generalized} and KL divergence \cite{van2014renyi} exhibit poor performance across all data settings, especially at 25\% and 50\% seen data. Over 75\% seen data, these measures relatively perform well but still underperform as compared to other measures. It indicates that these measures failed to capture effective relation between query samples and support samples. On the other hand, jaccard distance and manhattan similarity demonstrate average F1-Score across all data splits. However, cosine and L2 distance manages to outperform all other measures. Specifically, at 75\% data split cosine exhibit 0.8403 F1-Score for MuRIL-base-17-languages. These measures manages to perform consistently for most configuration across all data splits. It indicates their effectiveness to capture their relationship between query and support vectors even at less seen data.
On 4-way 5-shot experimental setting, performance of most similarity measures improves significantly except for hamming distance and KL divergence. They do not perform any improvement and still exhibit poor performance even with five examples per class. Moreover, bhattacharyya distance and chebyshev distance demonstrate F1-Score less than 45\% even at 75\% seen data. Even though jaccard, manhattan and dice similarity perform comparatively well but did not achieve promising results. However, euclidean emerges as one of the top performer across most of language models, indicates a significant improvement. Moreover, dot product and pearson correlation shows signifcant improvement and also among the top performer. However, these measures do not outperform cosine similarity. As cosine similarity shows consistent performance. 
Overall, cosine similarity is the most reliable and versatile similarity measure across various data splits and language models. The consistent performance across different configurations highlights its robustness.

%%%%%%%%%%%%%%%%%%%%%%%%%%%%%%%%%%%%%%%%%%%%%%%%%%%%%%%%%%%%%%%%%%%%%%%%%%%%
\begin{table}[htbp]
\caption{F1-Score of 4-way 1-shot and 4-way 5-shot over Web Queries dataset}
\label{Web_Sim}
\renewcommand{\arraystretch}{1.5}
\resizebox{1.0\textwidth}{!}{
\begin{tabular}{|l|c|cccccc|cccccc|}
\hline
\multicolumn{2}{|c|}{\textbf{Experimental Setting}} & \multicolumn{6}{c|}{\textbf{4-way 1-shot}} & \multicolumn{6}{c|}{\textbf{4-way 5-shot}} \\ \hline
\textbf{Model} & {\textbf{Seen Classes}} & \multicolumn{1}{l|}{\textbf{\begin{tabular}[c]{@{}l@{}}BERT-\\ base-\\ 15-\\ languages\end{tabular}}} & \multicolumn{1}{c|}{\textbf{\begin{tabular}[c]{@{}l@{}}BERT-\\ base-\\ 104-\\ languages\end{tabular}}} & \multicolumn{1}{c|}{\textbf{\begin{tabular}[c]{@{}l@{}}DistilBERT-\\ base-\\ 104-\\ languages\end{tabular}}} & \multicolumn{1}{c|}{\textbf{\begin{tabular}[c]{@{}l@{}}DeBERTa-\\ base-\\ 100-\\ languages\end{tabular}}} & \multicolumn{1}{c|}{\textbf{\begin{tabular}[c]{@{}l@{}}MuRIL-\\ base-\\ 100-\\ languages\end{tabular}}} & \textbf{\begin{tabular}[c]{@{}l@{}}RoBERTa-\\ small-\\ Urdu\end{tabular}} & \multicolumn{1}{c|}{\textbf{\begin{tabular}[c]{@{}l@{}}BERT-\\ base-\\ 15-\\ languages\end{tabular}}} & \multicolumn{1}{c|}{\textbf{\begin{tabular}[c]{@{}l@{}}BERT-\\ base-\\ 104-\\ languages\end{tabular}}} & \multicolumn{1}{c|}{\textbf{\begin{tabular}[c]{@{}l@{}}DistilBERT-\\ base-\\ 104-\\ languages\end{tabular}}} & \multicolumn{1}{c|}{\textbf{\begin{tabular}[c]{@{}l@{}}DeBERTa-\\ base-\\ 100-\\ languages\end{tabular}}} & \multicolumn{1}{c|}{\textbf{\begin{tabular}[c]{@{}l@{}}MuRIL-\\ base-\\ 100-\\ languages\end{tabular}}} & \textbf{\begin{tabular}[c]{@{}l@{}}RoBERTa-\\ small-\\ Urdu\end{tabular}} \\ \hline
\multirow{3}{*}{\begin{tabular}[c]{@{}l@{}}Angular\\ Similarity\end{tabular}} & 25 & \multicolumn{1}{l|}{0.2050} & \multicolumn{1}{l|}{0.2024} & \multicolumn{1}{l|}{0.2223} & \multicolumn{1}{l|}{0.2236} & \multicolumn{1}{l|}{\textbf{0.2893}} & 0.1645 & \multicolumn{1}{l|}{0.2041} & \multicolumn{1}{l|}{0.3058} & \multicolumn{1}{l|}{0.2356} & \multicolumn{1}{l|}{0.2464} & \multicolumn{1}{l|}{0.3350} & 0.2662 \\ \cline{2-14} 
 & 50 & \multicolumn{1}{l|}{0.3428} & \multicolumn{1}{l|}{0.3902} & \multicolumn{1}{l|}{0.2761} & \multicolumn{1}{l|}{0.2333} & \multicolumn{1}{l|}{0.3801} & 0.3018 & \multicolumn{1}{l|}{0.3690} & \multicolumn{1}{l|}{0.4576} & \multicolumn{1}{l|}{0.3532} & \multicolumn{1}{l|}{0.3546} & \multicolumn{1}{l|}{0.4783} & 0.3254 \\ \cline{2-14} 
 & 75 & \multicolumn{1}{l|}{0.6047} & \multicolumn{1}{l|}{0.6512} & \multicolumn{1}{l|}{0.5395} & \multicolumn{1}{l|}{0.4979} & \multicolumn{1}{l|}{0.7911} & 0.6574 & \multicolumn{1}{l|}{0.5529} & \multicolumn{1}{l|}{0.7055} & \multicolumn{1}{l|}{0.6427} & \multicolumn{1}{l|}{0.5628} & \multicolumn{1}{l|}{0.7689} & 0.6647 \\ \hline
\multirow{3}{*}{\begin{tabular}[c]{@{}l@{}}Bhattacharyya\\ Distance\end{tabular}} & 25 & \multicolumn{1}{l|}{0.0857} & \multicolumn{1}{l|}{0.0782} & \multicolumn{1}{l|}{0.1041} & \multicolumn{1}{l|}{0.0607} & \multicolumn{1}{l|}{0.1350} & 0.0770 & \multicolumn{1}{l|}{0.0684} & \multicolumn{1}{l|}{0.0654} & \multicolumn{1}{l|}{0.0749} & \multicolumn{1}{l|}{0.0843} & \multicolumn{1}{l|}{0.0992} & 0.1126 \\ \cline{2-14} 
 & 50 & \multicolumn{1}{l|}{0.1747} & \multicolumn{1}{l|}{0.1898} & \multicolumn{1}{l|}{0.1787} & \multicolumn{1}{l|}{0.1632} & \multicolumn{1}{l|}{0.2574} & 0.1297 & \multicolumn{1}{l|}{0.1502} & \multicolumn{1}{l|}{0.2452} & \multicolumn{1}{l|}{0.2577} & \multicolumn{1}{l|}{0.1432} & \multicolumn{1}{l|}{0.2845} & 0.1423 \\ \cline{2-14} 
 & 75 & \multicolumn{1}{l|}{0.3959} & \multicolumn{1}{l|}{0.3023} & \multicolumn{1}{l|}{0.4132} & \multicolumn{1}{l|}{0.3265} & \multicolumn{1}{l|}{0.3215} & 0.3621 & \multicolumn{1}{l|}{0.3899} & \multicolumn{1}{l|}{0.4332} & \multicolumn{1}{l|}{0.3918} & \multicolumn{1}{l|}{0.3771} & \multicolumn{1}{l|}{0.4515} & 0.3276 \\ \hline
\multirow{3}{*}{\begin{tabular}[c]{@{}l@{}}Chebyshev\\ Distance\end{tabular}} & 25 & \multicolumn{1}{l|}{0.0378} & \multicolumn{1}{l|}{0.0357} & \multicolumn{1}{l|}{0.0357} & \multicolumn{1}{l|}{0.0357} & \multicolumn{1}{l|}{0.0357} & 0.0385 & \multicolumn{1}{l|}{0.0414} & \multicolumn{1}{l|}{0.0478} & \multicolumn{1}{l|}{0.0357} & \multicolumn{1}{l|}{0.0367} & \multicolumn{1}{l|}{0.0357} & 0.0357 \\ \cline{2-14} 
 & 50 & \multicolumn{1}{l|}{0.0488} & \multicolumn{1}{l|}{0.0476} & \multicolumn{1}{l|}{0.0476} & \multicolumn{1}{l|}{0.0476} & \multicolumn{1}{l|}{0.0476} & 0.0486 & \multicolumn{1}{l|}{0.0476} & \multicolumn{1}{l|}{0.0476} & \multicolumn{1}{l|}{0.0476} & \multicolumn{1}{l|}{0.0485} & \multicolumn{1}{l|}{0.0476} & 0.0475 \\ \cline{2-14} 
 & 75 & \multicolumn{1}{l|}{0.1667} & \multicolumn{1}{l|}{0.1668} & \multicolumn{1}{l|}{0.1667} & \multicolumn{1}{l|}{0.1667} & \multicolumn{1}{l|}{0.1667} & 0.1667 & \multicolumn{1}{l|}{0.1690} & \multicolumn{1}{l|}{0.1667} & \multicolumn{1}{l|}{0.1667} & \multicolumn{1}{l|}{0.1667} & \multicolumn{1}{l|}{0.1667} & 0.1667 \\ \hline
\multirow{3}{*}{\begin{tabular}[c]{@{}l@{}}Cosine\\ Similarity\end{tabular}} & 25 & \multicolumn{1}{l|}{0.2290} & \multicolumn{1}{l|}{0.2188} & \multicolumn{1}{l|}{\textbf{0.2444}} & \multicolumn{1}{l|}{0.2333} & \multicolumn{1}{l|}{0.2867} & 0.2049 & \multicolumn{1}{l|}{0.3005} & \multicolumn{1}{l|}{0.3761} & \multicolumn{1}{l|}{0.2917} & \multicolumn{1}{l|}{0.2467} & \multicolumn{1}{l|}{0.3395} & 0.2892 \\ \cline{2-14} 
 & 50 & \multicolumn{1}{l|}{0.3605} & \multicolumn{1}{l|}{0.4279} & \multicolumn{1}{l|}{0.3598} & \multicolumn{1}{l|}{0.3009} & \multicolumn{1}{l|}{0.3614} & \textbf{0.3556} & \multicolumn{1}{l|}{\textbf{0.4128}} & \multicolumn{1}{l|}{\textbf{0.5130}} & \multicolumn{1}{l|}{0.4367} & \multicolumn{1}{l|}{\textbf{0.4058}} & \multicolumn{1}{l|}{0.5222} & \textbf{0.3785} \\ \cline{2-14} 
 & 75 & \multicolumn{1}{l|}{0.6567} & \multicolumn{1}{l|}{0.6962} & \multicolumn{1}{l|}{0.5622} & \multicolumn{1}{l|}{\textbf{0.5837}} & \multicolumn{1}{l|}{\textbf{0.8403}} & 0.6766 & \multicolumn{1}{l|}{0.7078} & \multicolumn{1}{l|}{0.7566} & \multicolumn{1}{l|}{0.7177} & \multicolumn{1}{l|}{\textbf{0.6716}} & \multicolumn{1}{l|}{\textbf{0.8442}} & 0.6933 \\ \hline
\multirow{3}{*}{\begin{tabular}[c]{@{}l@{}}Dice \\ Similarity\end{tabular}} & 25 & \multicolumn{1}{l|}{0.1637} & \multicolumn{1}{l|}{0.1276} & \multicolumn{1}{l|}{0.1644} & \multicolumn{1}{l|}{0.1869} & \multicolumn{1}{l|}{0.2309} & 0.2075 & \multicolumn{1}{l|}{0.1883} & \multicolumn{1}{l|}{0.1989} & \multicolumn{1}{l|}{0.2173} & \multicolumn{1}{l|}{0.2074} & \multicolumn{1}{l|}{0.2482} & 0.2342 \\ \cline{2-14} 
 & 50 & \multicolumn{1}{l|}{0.3202} & \multicolumn{1}{l|}{0.3826} & \multicolumn{1}{l|}{0.2820} & \multicolumn{1}{l|}{0.2250} & \multicolumn{1}{l|}{0.3320} & 0.2449 & \multicolumn{1}{l|}{0.3542} & \multicolumn{1}{l|}{0.3989} & \multicolumn{1}{l|}{0.3979} & \multicolumn{1}{l|}{0.3427} & \multicolumn{1}{l|}{0.3862} & 0.2689 \\ \cline{2-14} 
 & 75 & \multicolumn{1}{l|}{0.5915} & \multicolumn{1}{l|}{0.6563} & \multicolumn{1}{l|}{0.5216} & \multicolumn{1}{l|}{0.5386} & \multicolumn{1}{l|}{0.8018} & 0.6626 & \multicolumn{1}{l|}{0.5684} & \multicolumn{1}{l|}{0.6958} & \multicolumn{1}{l|}{0.6491} & \multicolumn{1}{l|}{0.6695} & \multicolumn{1}{l|}{0.7347} & 0.6313 \\ \hline
\multirow{3}{*}{\begin{tabular}[c]{@{}l@{}}Dot product\\ Similarity\end{tabular}} & 25 & \multicolumn{1}{l|}{\textbf{0.2455}} & \multicolumn{1}{l|}{0.2331} & \multicolumn{1}{l|}{0.2249} & \multicolumn{1}{l|}{0.2065} & \multicolumn{1}{l|}{0.2844} & 0.1584 & \multicolumn{1}{l|}{0.2310} & \multicolumn{1}{l|}{0.2370} & \multicolumn{1}{l|}{0.2828} & \multicolumn{1}{l|}{\textbf{0.2679}} & \multicolumn{1}{l|}{\textbf{0.3603}} & 0.2351 \\ \cline{2-14} 
 & 50 & \multicolumn{1}{l|}{0.3228} & \multicolumn{1}{l|}{0.4267} & \multicolumn{1}{l|}{0.3099} & \multicolumn{1}{l|}{0.1694} & \multicolumn{1}{l|}{0.3500} & 0.3008 & \multicolumn{1}{l|}{0.3381} & \multicolumn{1}{l|}{0.4986} & \multicolumn{1}{l|}{\textbf{0.4482}} & \multicolumn{1}{l|}{0.3758} & \multicolumn{1}{l|}{\textbf{0.5600}} & 0.3344 \\ \cline{2-14} 
 & 75 & \multicolumn{1}{l|}{0.6747} & \multicolumn{1}{l|}{0.6821} & \multicolumn{1}{l|}{0.5408} & \multicolumn{1}{l|}{0.4595} & \multicolumn{1}{l|}{0.7890} & 0.6667 & \multicolumn{1}{l|}{0.7464} & \multicolumn{1}{l|}{0.7625} & \multicolumn{1}{l|}{0.7050} & \multicolumn{1}{l|}{0.6411} & \multicolumn{1}{l|}{0.7695} & 0.6878 \\ \hline
\multirow{3}{*}{\begin{tabular}[c]{@{}l@{}}Euclidean\_\\ Distance\end{tabular}} & 25 & \multicolumn{1}{l|}{0.2141} & \multicolumn{1}{l|}{\textbf{0.2783}} & \multicolumn{1}{l|}{0.2356} & \multicolumn{1}{l|}{0.2019} & \multicolumn{1}{l|}{0.2492} & 0.2164 & \multicolumn{1}{l|}{0.2416} & \multicolumn{1}{l|}{0.3307} & \multicolumn{1}{l|}{0.3422} & \multicolumn{1}{l|}{0.2210} & \multicolumn{1}{l|}{0.3177} & 0.2426 \\ \cline{2-14} 
 & 50 & \multicolumn{1}{l|}{0.3255} & \multicolumn{1}{l|}{0.4409} & \multicolumn{1}{l|}{0.2794} & \multicolumn{1}{l|}{\textbf{0.3131}} & \multicolumn{1}{l|}{0.3115} & 0.2831 & \multicolumn{1}{l|}{0.3969} & \multicolumn{1}{l|}{0.4453} & \multicolumn{1}{l|}{0.4044} & \multicolumn{1}{l|}{0.3792} & \multicolumn{1}{l|}{0.4556} & 0.2996 \\ \cline{2-14} 
 & 75 & \multicolumn{1}{l|}{0.6742} & \multicolumn{1}{l|}{0.7045} & \multicolumn{1}{l|}{0.5437} & \multicolumn{1}{l|}{0.3562} & \multicolumn{1}{l|}{0.7852} & 0.5775 & \multicolumn{1}{l|}{0.6550} & \multicolumn{1}{l|}{0.7395} & \multicolumn{1}{l|}{0.7387} & \multicolumn{1}{l|}{0.6564} & \multicolumn{1}{l|}{0.8318} & 0.7069 \\ \hline
\multirow{3}{*}{\begin{tabular}[c]{@{}l@{}}Hamming\\ Distance\end{tabular}} & 25 & \multicolumn{1}{l|}{0.0357} & \multicolumn{1}{l|}{0.0357} & \multicolumn{1}{l|}{0.0357} & \multicolumn{1}{l|}{0.0357} & \multicolumn{1}{l|}{0.0357} & 0.0357 & \multicolumn{1}{l|}{0.0357} & \multicolumn{1}{l|}{0.0357} & \multicolumn{1}{l|}{0.0357} & \multicolumn{1}{l|}{0.0357} & \multicolumn{1}{l|}{0.0357} & 0.0357 \\ \cline{2-14} 
 & 50 & \multicolumn{1}{l|}{0.0476} & \multicolumn{1}{l|}{0.0476} & \multicolumn{1}{l|}{0.0476} & \multicolumn{1}{l|}{0.0476} & \multicolumn{1}{l|}{0.0476} & 0.0476 & \multicolumn{1}{l|}{0.0476} & \multicolumn{1}{l|}{0.0476} & \multicolumn{1}{l|}{0.0476} & \multicolumn{1}{l|}{0.0476} & \multicolumn{1}{l|}{0.0476} & 0.0476 \\ \cline{2-14} 
 & 75 & \multicolumn{1}{l|}{0.1667} & \multicolumn{1}{l|}{0.1667} & \multicolumn{1}{l|}{0.1667} & \multicolumn{1}{l|}{0.1667} & \multicolumn{1}{l|}{0.1667} & 0.1667 & \multicolumn{1}{l|}{0.1667} & \multicolumn{1}{l|}{0.1667} & \multicolumn{1}{l|}{0.1667} & \multicolumn{1}{l|}{0.1667} & \multicolumn{1}{l|}{0.1667} & 0.1667 \\ \hline
\multirow{3}{*}{\begin{tabular}[c]{@{}l@{}}Jaccard\\ Similarity\end{tabular}} & 25 & \multicolumn{1}{l|}{0.1589} & \multicolumn{1}{l|}{0.1657} & \multicolumn{1}{l|}{0.1937} & \multicolumn{1}{l|}{0.1922} & \multicolumn{1}{l|}{0.2492} & 0.1836 & \multicolumn{1}{l|}{0.2166} & \multicolumn{1}{l|}{0.1986} & \multicolumn{1}{l|}{0.2105} & \multicolumn{1}{l|}{0.2028} & \multicolumn{1}{l|}{0.2770} & 0.1768 \\ \cline{2-14} 
 & 50 & \multicolumn{1}{l|}{0.3116} & \multicolumn{1}{l|}{0.2866} & \multicolumn{1}{l|}{0.2223} & \multicolumn{1}{l|}{0.2198} & \multicolumn{1}{l|}{0.2967} & 0.2108 & \multicolumn{1}{l|}{0.2980} & \multicolumn{1}{l|}{0.3399} & \multicolumn{1}{l|}{0.3078} & \multicolumn{1}{l|}{0.3210} & \multicolumn{1}{l|}{0.3047} & 0.2993 \\ \cline{2-14} 
 & 75 & \multicolumn{1}{l|}{0.5670} & \multicolumn{1}{l|}{0.5717} & \multicolumn{1}{l|}{0.5229} & \multicolumn{1}{l|}{0.3548} & \multicolumn{1}{l|}{0.7341} & \textbf{0.7227} & \multicolumn{1}{l|}{0.5948} & \multicolumn{1}{l|}{0.6455} & \multicolumn{1}{l|}{0.6351} & \multicolumn{1}{l|}{0.6028} & \multicolumn{1}{l|}{0.7516} & 0.7032 \\ \hline
\multirow{3}{*}{\begin{tabular}[c]{@{}l@{}}KL\\ Divergence\end{tabular}} & 25 & \multicolumn{1}{l|}{0.0357} & \multicolumn{1}{l|}{0.0357} & \multicolumn{1}{l|}{0.0357} & \multicolumn{1}{l|}{0.0357} & \multicolumn{1}{l|}{0.0357} & 0.0357 & \multicolumn{1}{l|}{0.0357} & \multicolumn{1}{l|}{0.0357} & \multicolumn{1}{l|}{0.0357} & \multicolumn{1}{l|}{0.0357} & \multicolumn{1}{l|}{0.0357} & 0.0357 \\ \cline{2-14} 
 & 50 & \multicolumn{1}{l|}{0.0476} & \multicolumn{1}{l|}{0.0476} & \multicolumn{1}{l|}{0.0476} & \multicolumn{1}{l|}{0.0476} & \multicolumn{1}{l|}{0.0476} & 0.0476 & \multicolumn{1}{l|}{0.0476} & \multicolumn{1}{l|}{0.0476} & \multicolumn{1}{l|}{0.0476} & \multicolumn{1}{l|}{0.0476} & \multicolumn{1}{l|}{0.0476} & 0.0476 \\ \cline{2-14} 
 & 75 & \multicolumn{1}{l|}{0.1667} & \multicolumn{1}{l|}{0.1667} & \multicolumn{1}{l|}{0.1667} & \multicolumn{1}{l|}{0.1667} & \multicolumn{1}{l|}{0.1667} & 0.1667 & \multicolumn{1}{l|}{0.1667} & \multicolumn{1}{l|}{0.1667} & \multicolumn{1}{l|}{0.1667} & \multicolumn{1}{l|}{0.1667} & \multicolumn{1}{l|}{0.1667} & 0.1667 \\ \hline
\multirow{3}{*}{\begin{tabular}[c]{@{}l@{}}L2\\ Distance\end{tabular}} & 25 & \multicolumn{1}{l|}{0.2452} & \multicolumn{1}{l|}{0.1587} & \multicolumn{1}{l|}{0.2059} & \multicolumn{1}{l|}{0.1912} & \multicolumn{1}{l|}{0.2570} & \textbf{0.2245} & \multicolumn{1}{l|}{\textbf{0.3400}} & \multicolumn{1}{l|}{0.2168} & \multicolumn{1}{l|}{\textbf{0.3342}} & \multicolumn{1}{l|}{0.2211} & \multicolumn{1}{l|}{0.3253} & 0.2364 \\ \cline{2-14} 
 & 50 & \multicolumn{1}{l|}{\textbf{0.3677}} & \multicolumn{1}{l|}{\textbf{0.4471}} & \multicolumn{1}{l|}{\textbf{0.3700}} & \multicolumn{1}{l|}{0.3000} & \multicolumn{1}{l|}{\textbf{0.3696}} & 0.2171 & \multicolumn{1}{l|}{0.3262} & \multicolumn{1}{l|}{0.4585} & \multicolumn{1}{l|}{0.4304} & \multicolumn{1}{l|}{0.3726} & \multicolumn{1}{l|}{0.4646} & 0.3022 \\ \cline{2-14} 
 & 75 & \multicolumn{1}{l|}{\textbf{0.6925}} & \multicolumn{1}{l|}{0.5444} & \multicolumn{1}{l|}{\textbf{0.6391}} & \multicolumn{1}{l|}{0.5180} & \multicolumn{1}{l|}{0.7958} & 0.6796 & \multicolumn{1}{l|}{\textbf{0.7537}} & \multicolumn{1}{l|}{\textbf{0.7830}} & \multicolumn{1}{l|}{\textbf{0.7379}} & \multicolumn{1}{l|}{0.5611} & \multicolumn{1}{l|}{0.8261} & \textbf{0.7451} \\ \hline
\multirow{3}{*}{\begin{tabular}[c]{@{}l@{}}Manhattan\\ Distances\end{tabular}} & 25 & \multicolumn{1}{l|}{0.1480} & \multicolumn{1}{l|}{0.1434} & \multicolumn{1}{l|}{0.1371} & \multicolumn{1}{l|}{\textbf{0.2492}} & \multicolumn{1}{l|}{0.1961} & 0.1707 & \multicolumn{1}{l|}{0.3028} & \multicolumn{1}{l|}{0.2620} & \multicolumn{1}{l|}{0.2312} & \multicolumn{1}{l|}{0.2241} & \multicolumn{1}{l|}{0.2708} & 0.1766 \\ \cline{2-14} 
 & 50 & \multicolumn{1}{l|}{0.2831} & \multicolumn{1}{l|}{0.2916} & \multicolumn{1}{l|}{0.2233} & \multicolumn{1}{l|}{0.1494} & \multicolumn{1}{l|}{0.1992} & 0.1386 & \multicolumn{1}{l|}{0.4056} & \multicolumn{1}{l|}{0.3346} & \multicolumn{1}{l|}{0.2438} & \multicolumn{1}{l|}{0.2923} & \multicolumn{1}{l|}{0.3560} & 0.2703 \\ \cline{2-14} 
 & 75 & \multicolumn{1}{l|}{0.4633} & \multicolumn{1}{l|}{0.3310} & \multicolumn{1}{l|}{0.4450} & \multicolumn{1}{l|}{0.3608} & \multicolumn{1}{l|}{0.4489} & 0.6828 & \multicolumn{1}{l|}{0.6842} & \multicolumn{1}{l|}{0.5941} & \multicolumn{1}{l|}{0.5150} & \multicolumn{1}{l|}{0.5068} & \multicolumn{1}{l|}{0.8137} & 0.6066 \\ \hline
\multirow{3}{*}{\begin{tabular}[c]{@{}l@{}}Pearson\\ Correlation\end{tabular}} & 25 & \multicolumn{1}{l|}{0.2056} & \multicolumn{1}{l|}{0.1944} & \multicolumn{1}{l|}{0.2299} & \multicolumn{1}{l|}{0.1960} & \multicolumn{1}{l|}{0.2623} & 0.1916 & \multicolumn{1}{l|}{0.2406} & \multicolumn{1}{l|}{\textbf{0.3765}} & \multicolumn{1}{l|}{0.2348} & \multicolumn{1}{l|}{0.2634} & \multicolumn{1}{l|}{0.3515} & \textbf{0.2997} \\ \cline{2-14} 
 & 50 & \multicolumn{1}{l|}{0.3481} & \multicolumn{1}{l|}{0.4339} & \multicolumn{1}{l|}{0.3493} & \multicolumn{1}{l|}{0.2827} & \multicolumn{1}{l|}{0.3282} & 0.3252 & \multicolumn{1}{l|}{0.3810} & \multicolumn{1}{l|}{0.4728} & \multicolumn{1}{l|}{0.4229} & \multicolumn{1}{l|}{0.3423} & \multicolumn{1}{l|}{0.5262} & 0.3567 \\ \cline{2-14} 
 & 75 & \multicolumn{1}{l|}{0.6090} & \multicolumn{1}{l|}{0.6548} & \multicolumn{1}{l|}{0.5752} & \multicolumn{1}{l|}{0.5468} & \multicolumn{1}{l|}{0.8026} & 0.6488 & \multicolumn{1}{l|}{0.6825} & \multicolumn{1}{l|}{0.7631} & \multicolumn{1}{l|}{0.7027} & \multicolumn{1}{l|}{0.6528} & \multicolumn{1}{l|}{0.8366} & 0.6896 \\ \hline
\end{tabular}}
\end{table}
%%%%%%%%%%%%%%%%%%%%%%%%%%%%%%%%%%%%%%%%%%%%%%%%%%%%%%%%%%%%%%%%%%%%%%%%%%%%
\section{Proposed framework performance comparison with existing intent detection predictors}
To ensure a fair performance comparison of distinct predictive pipelines of the proposed framework with previous study \cite{shams2022improving}, we utilized the original Web Queries dataset containing three intent classes: Navigational, Transactional and Informational. Instead of experimenting with seen and unseen class settings, we conducted our experiments using the same three intent classes in both the training and testing sets. Following standard practices in existing studies \cite{shams2022improving} 80\% of the data is used for training and 20\% for testing. Furthermore, in 3-way 1-shot setting, one sample from each class is randomly selected as the support set, while remaining samples formed the query set for training the model. For testing, the trained model used one sample as the support set and predicted the intent of all other samples in the query set. Similarly, in the 3-way 5-shot setting, five random samples are selected from each class as the support set for both the training and testing phases. Table \ref{Comp} provides a detailed performance comparison between the existing predictors and the six distinct predictive pipelines of the proposed framework. 
Shams et al. \cite{shams2022improving} analyzed the impact of using different word vector models as input embedding layers for the U-IntentCapsNet predictor. Specifically, W2V-100 + U-IntentCapsNet, W2V-200 + U-IntentCapsNet, and W2V-300 + U-IntentCapsNet utilized 100, 200, and 300-dimensional Word2Vec embeddings, respectively. Additionally, mBERT embeddings were also used as input for the U-IntentCapsNet predictor. It is evident from Table \ref{Comp} that the performance of the model improves as the dimensionality of the Word2Vec embeddings increases. Moreover, the W2V-200 and W2V-300 pipelines outperformed the mBERT-based pipeline which demonstrates that the W2V-based baselines performed better than contextualized embeddings. However, U-IntentCapsNet emerged as the top performer among the existing predictors, significantly outperforming other models with a substantial margin.
%%%%%%%%%%%%%%%%%%%%%%%%%%%%%%%%%%%%%%%%%%%%%%%%%%%%
\begin{table}[htbp]
\centering
\caption{Performance comparison of distinct predictive pipelines of proposed framewrok with existing intent detection predictor}
\label{Comp}
\renewcommand{\arraystretch}{1.25}
\resizebox{1.0\textwidth}{!}{
\begin{tabular}{|c|l|c|l|l|l|l|}
\hline
\textbf{Author} & \textbf{Model} & \textbf{\begin{tabular}[c]{@{}l@{}}Experimental \\ Setting\end{tabular}} & \textbf{Accuracy} & \textbf{Precision} & \textbf{Recall} & \textbf{F1-Score} \\ \hline
\multirow{5}{*}{\rotatebox{90}{\begin{tabular}[c]{@{}l@{}}Shams et al. \\\cite{shams2022improving}\end{tabular}}} & \begin{tabular}[c]{@{}l@{}}W2V-100 + \\ U-IntentCapsNet\end{tabular} & \_ & 0.8242 & 0.8110 & 0.8242 & 0.8136 \\ \cline{2-7} 
 & \begin{tabular}[c]{@{}l@{}}W2V-200 + \\ U-IntentCapsNet\end{tabular} & \_ & 0.8561 & 0.8524 & 0.8561 & 0.8533 \\ \cline{2-7} 
 & \begin{tabular}[c]{@{}l@{}}W2V-300 + \\ U-IntentCapsNet\end{tabular} & \_ & 0.8613 & 0.8566 & 0.8613 & 0.8583 \\ \cline{2-7} 
 & \begin{tabular}[c]{@{}l@{}}mBERT + \\ U-IntentCapsNet\end{tabular} & \_ & 0.8394 & 0.8432 & 0.8394 & 0.8391 \\ \cline{2-7} 
 & U-IntentCapsNet & \_ & 0.9112 & 0.9083 & 0.9112 & 0.9084 \\ \hline
\multirow{6}{*}{\rotatebox{90}{\begin{tabular}[c]{@{}l@{}}Tahir et al. \\\cite{tahir2024benchmarking}\end{tabular}}} & GPT 3.5 & \multirow{6}{*}{0 shot} & \_ & \_ & \_ & 0.3000 \\ \cline{2-2} \cline{4-7} 
 & Bloomz 3B &  & \_ & \_ & \_ & 0.2200 \\ \cline{2-2} \cline{4-7} 
 & Bloomz 7B1 &  & \_ & \_ & \_ & 0.1800 \\ \cline{2-2} \cline{4-7} 
 & Llama 2 &  & \_ & \_ & \_ & 0.0700 \\ \cline{2-2} \cline{4-7} 
 & Llama 3.1 &  & \_ & \_ & \_ & 0.4200 \\ \cline{2-2} \cline{4-7} 
 & Ministral 8B &  & \_ & \_ & \_ & 0.3400 \\ \hline
\multirow{12}{*}{\rotatebox{90}{Pipelines of Proposed Framework}} & \multirow{2}{*}{\begin{tabular}[c]{@{}l@{}}bert-base-\\ 15-languages\end{tabular}} & 3-way 1-shot & 0.7655 & 0.8050 & 0.7655 & 0.7705 \\ \cline{3-7} 
 &  & 3-way 5-shot & 0.8208 & 0.8318 & 0.8208 & 0.8235 \\ \cline{2-7} 
 & \multirow{2}{*}{\begin{tabular}[c]{@{}l@{}}bert-base-\\ 104-languages\end{tabular}} & 3-way 1-shot & 0.7715 & 0.8047 & 0.7715 & 0.7771 \\ \cline{3-7} 
 &  & 3-way 5-shot & 0.8208 & 0.8352 & 0.8208 & 0.8245 \\ \cline{2-7} 
 & \multirow{2}{*}{\begin{tabular}[c]{@{}l@{}}distilbert-base-\\ 104-languages\end{tabular}} & 3-way 1-shot & 0.6453 & 0.7303 & 0.6453 & 0.6594 \\ \cline{3-7} 
 &  & 3-way 5-shot & 0.7923 & 0.7956 & 0.7923 & 0.7922 \\ \cline{2-7} 
 & \multirow{2}{*}{\begin{tabular}[c]{@{}l@{}}deberta-base-\\ 100-languages\end{tabular}} & 3-way 1-shot & 0.7976 & 0.8268 & 0.7976 & 0.8033 \\ \cline{3-7} 
 &  & 3-way 5-shot & 0.8310 & 0.8344 & 0.8310 & 0.8321 \\ \cline{2-7} 
 & \multirow{2}{*}{\begin{tabular}[c]{@{}l@{}}MuRIL-base-\\ 17-languages\end{tabular}} & 3-way 1-shot & 0.7536 & 0.8165 & 0.7536 & 0.7578 \\ \cline{3-7} 
 &  & 3-way 5-shot & 0.8778 & 0.8806 & 0.8778 & 0.8787 \\ \cline{2-7} 
 & \multirow{2}{*}{\begin{tabular}[c]{@{}l@{}}roberta-small-\\ Urdu\end{tabular}} & 3-way 1-shot & 0.9138 & 0.9299 & 0.9138 & 0.9164 \\ \cline{3-7} 
 &  & 3-way 5-shot & 0.9552 & 0.9560 & 0.9552 & 0.9555 \\ \hline
\end{tabular}}
\end{table}
%%%%%%%%%%%%%%%%%%%%%%%%%%%%%%%%%%%%%%%%%%%%%%%%%
Tahir et al. \cite{tahir2024benchmarking} investigated the effectiveness of 6 different LLMs including GPT 3.5 \cite{ye2023comprehensive}, Bloomz 3B\cite{muennighoff2022crosslingual}, Bloomz 7B1\cite{muennighoff2022crosslingual}, Llama 2\cite{touvron2023llama}, Llama 3.1\cite{grattafiori2024llama} and Ministral 8B \cite{mistral2024ministraux} in zero-shot setting by prompt optimization. Prompt optimization involved as to classify the given query in one of three predefined labels. As shown in Table \ref{Comp}, all the LLMs remain failed to capture disciminative features among different classes. Specifically, Llama 2 demonstrate lowest F1-Score while Llama 3.1 outperform other LLMs in zero-shot setting followed by Ministral 8B and GPT 3.5. However, these models did not achieve significantly results.
On the other hand, predictive pipelines of proposed framework
Under the 3-way 1-shot setting, models such as Muril-Base-17-Languages, BERT-Base-15-Languages, BERT-Base-104-Languages, and DistilBERT-Base-104-Languages performed the worst, with F1 scores remaining below 0.80. However, performance of all 4 predictors improved slightly under the 4-way 5-shot setting but, Muril-base-17-Languages demonstrated a notable performance increase of 12\%. In contrast, Roberta-Small-Urdu significantly outperformed other predictor and achieved an impressive F1-Score of 0.91 in the 3-way 1-shot setting and 0.95 in the 3-way 5-shot setting. Hence the top performer predictor Roberta-Small-Urdu of proposed framework outperforms existing top performing predictor with a significant margin of around 4\% in F1-Score. 
It is evident from Table \ref{Comp} that RoBERTa-small-Urdu pipeline achieved superior performance compared to Shams et al. \cite{shams2022improving} predictor with a significant margin of 4.71\% in terms of F1-Score. Shams et al., 2022 \cite{shams2022improving} leverages BiLSTM with Capsule Neural Networks for intent detection to preserve hierarchical relationships between features. However, proposed framework utilizes pre-trained language models, which provide richer contextual embeddings. Additionally, re-train these models on domain-specific data, allows better adaptation to the target domain. This combination results in superior accuracy compared to the BiLSTM-Capsule approach, as it effectively captures both general linguistic knowledge and domain-specific nuances. Moreover, it also highlights that the size of support set influences model performance, as 5-shot setting offer improved results compared to the 1-shot setting due to the availability of additional context. Overall, the findings underline the flexibility and robustness of the proposed framework, making it a promising approach for intent classification in few-shot learning scenarios. Moreover, the predictive pipelines were constructed using re-trained language models enabled the pipelines to better capture the nuances of the three intent classes.

\normalsize
\section{Conclusion}
This research empowers Urdu intent detection by developing a robust and accurate large-scale framework encompassing distinct predictive pipelines. Framework facilitates development of end-to-end Urdu intent detection predictive pipelines by leveraging unique contrastive learning strategy and a prototype-informed attention mechanism. A deep analysis of framework predictive pipelines performance across 2 datasets reveals that contrastive learning based re-training strategy significantly enhance the performance of all 6 language models. In addition, within framework predictive pipelines, among six distinct language models across two Urdu intent detection datasets reveals that predictive pipelines produce better performance by utilizing multi-lingual language models in comparison to Urdu language specific language model in few-shot setting. Furthermore, within diverse types of few-shot-learning based Urdu intent detection predictive pipelines, a large-scale analysis of 13 similarity computation methods reveals that cosine method utilization in the predictive pipeline is appropriate to develop a better application along with MuRIL language model. Conversely, in traditional approaches, RoBERTa-small-Urdu demonstrates strong performance.

The findings of this study contribute significantly to the advancement of Urdu language processing, specifically in the critical area of intent detection, and provide a solid foundation for future research aimed at developing more sophisticated and personalized AI-driven interactions for Urdu-speaking users. A compiling future direction of this work is to train diverse types of language models on on large Urdu language data and use these models in an end-to-end intent detection predictive pipeline. In addition, Urdu intent detection landscape, also lacks the potential exploration of advanced generative language models including DeepSeek, Gemini and GPT.


% %% For authoryear reference style

% \bibitem[Lamport(1994)]{lamport94}
%   \textit{\LaTeX: a document preparation system},
%   Addison Wesley, Massachusetts,

% \end{thebibliography}

\bibliographystyle{elsarticle-harv}   % Style of the references (Harvard style in this case)
\bibliography{References} 
\end{document}